\documentclass[sigconf]{acmart}
\AtBeginDocument{%
  }

\usepackage[ruled,vlined]{algorithm2e}
\usepackage{amsmath}
\usepackage{float}
\usepackage{tcolorbox}
\usepackage{enumitem}
\usepackage{xcolor}
\usepackage{graphicx}
\usepackage{booktabs}
\usepackage{balance}
\usepackage{xspace}

\copyrightyear{2026}
\acmYear{2026}
\setcopyright{cc}
\setcctype{by}
\acmConference[KDD '26]{Proceedings of the 32nd ACM SIGKDD Conference on Knowledge Discovery and Data Mining V.2}{August 09--13, 2026}{Jeju Island, Republic of Korea}
\acmBooktitle{Proceedings of the 32nd ACM SIGKDD Conference on Knowledge Discovery and Data Mining V.2 (KDD '26), August 09--13, 2026, Jeju Island, Republic of Korea}
\acmDOI{10.1145/3770855.3818202}
\acmISBN{979-8-4007-2259-2/2026/08}

\begin{document}

\title[LatentCRS: A Variational EM Framework for Bridging Semantics and Behavior in CRS]{LatentCRS: A Variational EM Framework for Bridging Semantics and Behavior in LLM-based Conversational Recommendation}
\author{Guanrong Li}
\affiliation{%
  \institution{Nanjing University}
  \city{Nanjing}
  \state{Jiangsu}
  \country{China}
}
\email{grli@smail.nju.edu.cn}

\author{Kuo Tian}
\affiliation{%
  \institution{Nanjing University}
  \city{Nanjing}
  \state{Jiangsu}
  \country{China}
}
\email{tiank@smail.nju.edu.cn}

\author{Jinnan Qi}
\affiliation{%
  \institution{Nanjing University}
  \city{Nanjing}
  \state{Jiangsu}
  \country{China}
}
\email{qijinnan@smail.nju.edu.cn}

\author{Qinghan Fu}
\affiliation{%
  \institution{Nanjing University}
  \city{Nanjing}
  \state{Jiangsu}
  \country{China}
}
\email{njuqhfu@gmail.com}

\author{Zhen Wu}
\authornote{Corresponding Author}
\affiliation{%
  \institution{Nanjing University}
  \city{Nanjing}
  \state{Jiangsu}
  \country{China}
}
\email{wuz@nju.edu.cn}

\author{Rui Xia}
\affiliation{%
  \institution{Nanjing University}
  \city{Nanjing}
  \state{Jiangsu}
  \country{China}
}
\email{rxia@nju.edu.cn}

\author{Xinyu Dai}
\affiliation{%
  \institution{Nanjing University}
  \city{Nanjing}
  \state{Jiangsu}
  \country{China}
}
\email{daixinyu@nju.edu.cn}

\renewcommand{\shortauthors}{Guanrong Li et al.}
\newcommand\ours{\textsc{LatentCRS}\xspace}

\begin{abstract}
Conversational Recommender Systems (CRS) powered by Large Language Models (LLMs) enable users to articulate explicit and dynamic preferences, overcoming the limitations of fixed templates. However, despite their superior semantic proficiency, LLMs have not yet achieved corresponding improvements in recommendation accuracy. This discrepancy arises from a fundamental representation gap: while LLMs operate within a semantic space, they lack the behavioral grounding needed to encode user behavioral patterns, such as item co-occurrences, which are crucial for accurate recommendations. To address this, we propose a model-agnostic Variational EM Framework for Bridging Semantics and Behavior in LLM-based Conversational Recommendation (\ours). Based on the observation that dialogue and interactions reflect the same latent intent, \ours uses a variational expectation-maximization (EM) procedure, where user intent connects semantic representations with behavioral patterns. Extensive experiments on real-world datasets demonstrate that \ours effectively bridges the representation gap and outperforms baselines.
\end{abstract}




\maketitle

\begin{figure}[t]
  \includegraphics[width=0.7\linewidth]{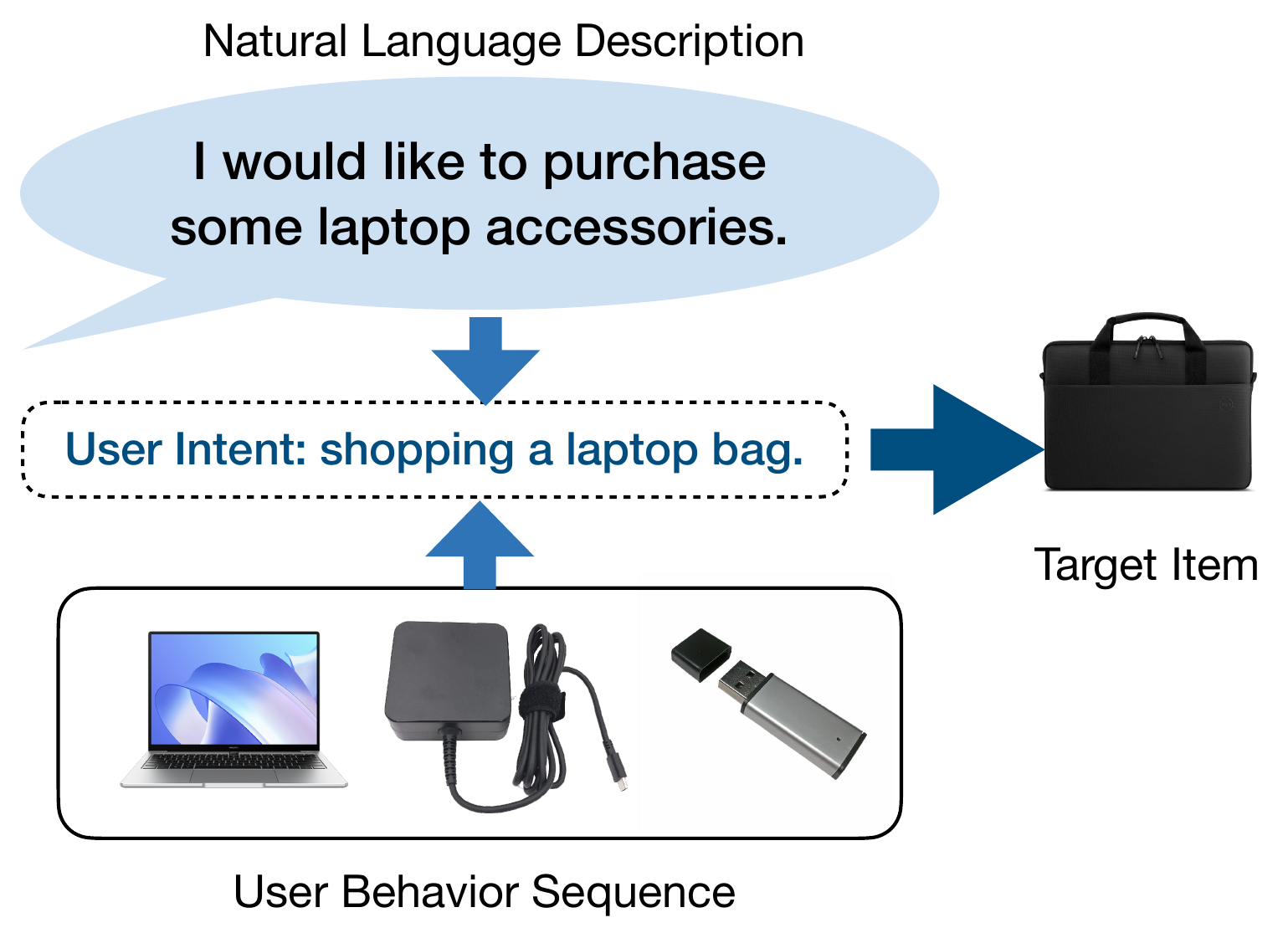}
  \caption{User intent inferred from behavior sequences and natural language descriptions. This demonstrates that both natural language descriptions and user behavior sequences serve as observations of latent user intent.}
  \label{fig:intent}
\end{figure}

\section{Introduction}\label{intro}
Conversational recommender systems (CRS) deliver personalized suggestions by engaging users in active dialogue, allowing users to explicitly describe and refine their interests through multi-turn interactions, enhancing recommendation accuracy and improving user experience \cite{siro2023understanding, mahmud2025understanding}.
Transcending the rigid templates of early approaches \cite{SunZ18, zhang2018towards}, recent Large Language Models (LLMs) have revolutionized the domain by enabling natural, fluid conversations where users can articulate dynamic preferences far beyond the limitations of predefined rules \cite{yang2024unleashing, gao2023chat}.

Despite advances in the linguistic capabilities of LLMs \cite{wei2022chain, luo2023wizardcoder}, their superior semantic proficiency has not yet translated into commensurate recommendation accuracy.
Accurate conversational recommendation necessitates the synergy of two distinct feature spaces: the semantic space, which represents explicit user demands extracted from dialogue, and the behavioral space, which captures interaction patterns derived from user history. However, LLMs trained solely on text lack the intrinsic capability to capture non-linguistic behavioral correlations. For instance, the classic ``beer and diapers'' observation illustrates a frequent co-purchase pattern with no inherent semantic relationship, a dependency that purely text-based LLMs struggle to encode \cite{liu2023chatgpt}.

To bridge these disparate representation spaces, existing methods adopt an agent-based framework, treating the recommendation model as a tool \cite{gao2023chat, huang2023recommender}. However, this architecture fundamentally decouples dialogue understanding from the recommendation process. As the recommendation model operates in isolation, it cannot directly attend to the specific preferences active in the conversation. Consequently, the lack of dialogue context limits ranking accuracy.
Alternatively, some approaches aim to unify semantic and behavioral spaces by converting items into textual descriptions and fine-tuning LLMs on user behavior data. However, these methods face scalability issues due to the LLMs' context window, which limits the ability to process large item catalogs. Additionally, fine-tuning on extensive user data is computationally expensive, making it impractical for real-world use. Thus, how to efficiently integrate semantic and behavioral spaces for dialogue-aware recommendations remains a key challenge.

To address these challenges, we propose latent user intent as the shared conceptual bridge to close the gap between linguistic processing and behavioral modeling. 
Our approach is grounded in the insight that linguistic expressions and behavioral patterns are not independent entities; rather, they are distinct surface manifestations of a single underlying user intent. For instance, as illustrated in Figure~\ref{fig:intent}, a user's latent intent to purchase a laptop bag may be explicitly articulated through the utterance, ``I would like to purchase some laptop accessories'' while simultaneously being implied by a behavioral sequence, such as purchasing a laptop and peripherals without a corresponding case. 
This implies that latent intent acts as the unifying variable, connecting the implicit patterns derived from behavior sequences with the explicit interests expressed through dialogue.
We propose a Variational EM Framework for Bridging Semantics and Behavior in LLM-based Conversational Recommendation (\ours), a model-agnostic framework designed to bridge the representation gap through latent user intent. 
Recognizing that intent is not directly observable, we treat it as a hidden variable governing both modalities. \ours employs a variational expectation–maximization (EM) procedure: an inference module estimates the intent distribution from behavior, while a generative module produces recommendations conditioned on this estimated intent. This alternating optimization effectively harmonizes semantic and behavioral representations with a theoretically grounded optimization objective. Crucially, \ours operates as a parameter-efficient solution, enabling seamless integration with diverse LLMs and recommendation models without resource-intensive fine-tuning.

Extensive experiments demonstrate that \ours significantly outperforms state-of-the-art baselines in both single-turn and multi-turn settings. The main contributions are summarized as follows:

\begin{itemize}[leftmargin=*, nosep] 
\item We identify the fundamental representation gap between the linguistic semantic space of LLMs and the behavioral patterns required for recommendation. To address this, we propose a novel paradigm that utilizes latent intent as a shared conceptual bridge to unify these disparate representation spaces.
\item We propose \ours, a model-agnostic framework that bridges this gap via a variational EM procedure. Moreover, \ours operates as an efficient solution, enabling seamless integration with diverse LLMs and recommendation backbones without the need for resource-intensive parameter updates. 
\item Extensive experiments on multiple real-world datasets demonstrate that \ours consistently outperforms state-of-the-art baselines in both single-turn and multi-turn settings while requiring lower computational cost than LLM-based baselines.
\end{itemize}

\section{Related Work}
\subsection{Conversational Recommender Systems}
Conversational recommender systems (CRS) aim to provide accurate recommendations through interactions with users. Early approaches usually treated the recommendation and conversation components as two separate modules \cite{SunZ18, bi2019conversational, zhang2020conversational}. Initial work focused on how to interact with users by determining which attributes or items to query and by extracting information from user responses for more accurate recommendations \cite{christakopoulou2016towards, christakopoulou2018q}. Subsequent research extended these ideas to multi-turn conversational settings, often using reinforcement learning methods \cite{EstimationActionReflection2020, li2018towards, zhang2018towards, hu2022learning}. Although these methods were successful, they restricted user responses to binary signals (like or dislike), thereby limiting the information that users could provide.
The growing use of LLMs has introduced a new paradigm for interacting with users, which inspired the design of LLM-based CRS \cite{jin2023lending, gao2023chat, he2023large, huang2023recommender, bao2024large}. Some methods treat an LLM as an agent and the recommendation model as a tool \cite{gao2023chat, huang2023recommender, kemper2024retrieval}. Although this approach allows more expressive user responses, it still treats conversation and recommendation separately and only merges them through a binary flag to indicate when a recommendation should be made.
Other methods integrate recommendation and conversation in a single framework. Because LLMs have strong text processing and reasoning abilities, some systems rely solely on LLMs to perform both tasks. For instance, certain approaches follow the ReAct framework \cite{DBLP:conf/iclr/YaoZYDSN023}, where the LLM is instructed to plan and reason, carry out specific actions, and adjust if mistakes are detected \cite{gao2023chat, he2023large, he2024reindex, xi2024memocrs}. Further approaches incorporate additional training of LLMs using interaction data \cite{yang2024item, yang2024unleashing, ravaut2024parameter}. However, these strategies ignore the collaborative knowledge found in traditional recommender systems, and they can be expensive due to frequent LLM calls or extensive fine-tuning. Moreover, their performance can be uncertain because many studies only compare them with baselines in cold-start situations \cite{jin2023lending, yang2024unleashing}.
In contrast, we propose a more efficient approach that uses user intent as a bridge to incorporate users' long-term preferences from historical behavior and real-time interests from dialogue, enabling more accurate and adaptive recommendations and interactions with users.

\subsection{Intent Modeling in Recommendation}
Recently, many studies have examined user intent to improve recommendation accuracy \cite{cen2020controllable, liu2020basket, pan2020intent, tanjim2020attentive}. Most approaches focus on identifying common user intent. For example, MIND \cite{li2019multi} introduces a multi-interest extractor layer based on a capsule routing mechanism to handle multiple user intents. ISRec \cite{li2021intention} builds an intention graph to uncover each user's underlying intentions and employs a message-passing mechanism to predict future intentions. ICLRec \cite{chen2022intent} leverages an expectation-maximization framework to integrate learned intents into sequential recommendation models via contrastive self-supervised learning. Furthermore, PO4ISR \cite{sun2024large} incorporates intent modeling with LLMs for session recommendation, allowing these models to discern diverse user intents at a semantic level by iteratively refining and adjusting prompts. 
However, current intent modeling methods mostly concentrate on discovering similar user intents from behavioral data \cite{farshidi2024understanding}. In contrast, we argue that user intent can also act as a link between user behavior and interests expressed in natural language for CRS.

\begin{figure*}[t]
  \includegraphics[width=0.8\textwidth]{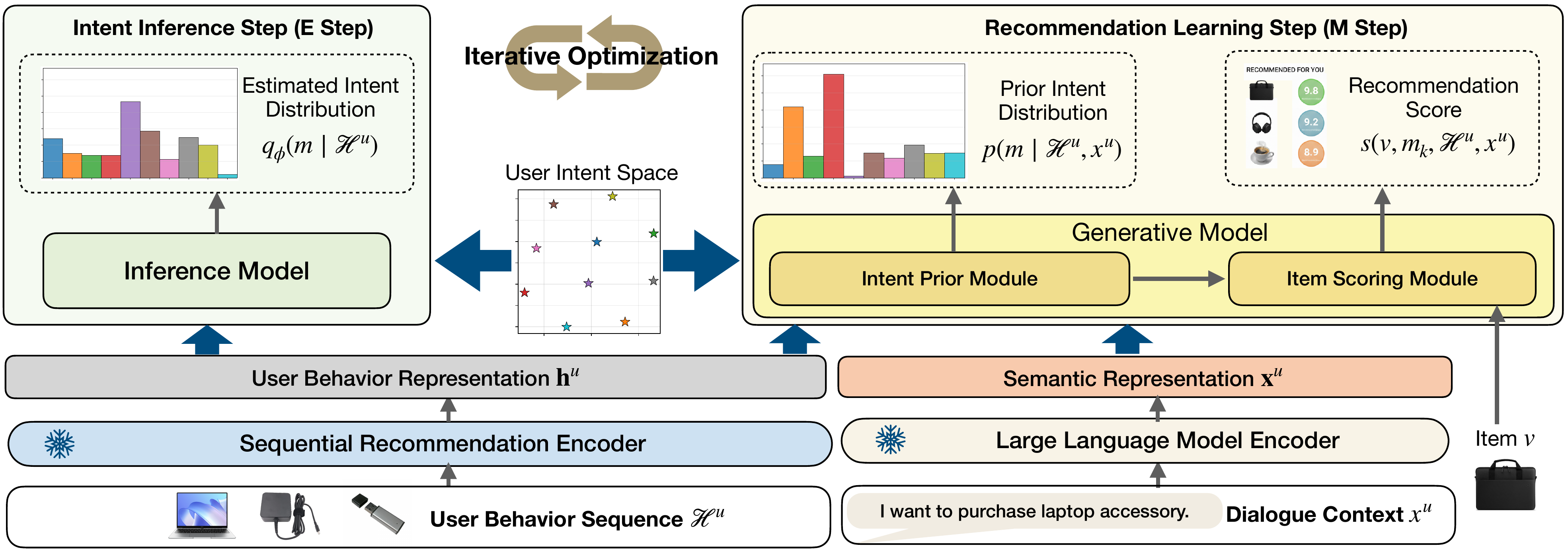}
  \caption{Overall structure of \ours. The left part represents the Intent Inference Step (E-step), which aims to accurately estimate the user's latent intent distribution $q_\phi(m \mid \mathcal{H}^u)$ with an inference model. The right part depicts the Recommendation Learning Step (M-step), where the generative model is optimized to align its intent prior $p_\theta(m \mid \mathcal{H}^u, x^u)$ with the inferred distribution and maximize the likelihood of the target item. To avoid the computationally prohibitive cost of normalizing over the full item space, we estimate an unnormalized recommendation score $s(v, m, \mathcal{H}^u, x^u)$ instead of exact probabilities.}
  \label{fig:structure}
\end{figure*}

\section{Methodology}
In this section, we present \ours, a model-agnostic framework designed to bridge the semantic-behavioral gap via latent user intent. We first establish the theoretical foundation by formulating the problem as a probabilistic latent variable model. We then detail the proposed variational EM procedure, which alternates between an Intent Inference step (E-step) to estimate the underlying user intent distribution and a Recommendation Learning step (M-step) to generate intent-aware recommendations. Finally, we outline the optimization strategy used to ensure robust convergence.

\subsection{Conversational Recommendation with Latent User Intent}\label{intent_representation}
\subsubsection{\textbf{Probabilistic Formulation}}
Let $\mathcal{U}$ and $\mathcal{V}$ denote the set of users and items, respectively. For each user $u \in \mathcal{U}$, the historical interaction sequence is denoted as $\mathcal{H}^u = [v_1^u, \dots, v_{|\mathcal{H}^u|}^u]$, where $v_t^u \in \mathcal{V}$ represents the item interacted with at step $t$.
The $N$-turn conversational context is defined as a dialogue session $x^u = \{(q_j^u, r_j^u)\}_{j=1}^N$, consisting of user utterances $q_j^u$ and system responses $r_j^u$.
The goal of the CRS is to recommend the optimal target item $v$ given both the interaction history and dialogue context. We formulate this as a maximum likelihood estimation problem:
\begin{equation}\label{crs_raw}
\max_{\theta} \sum_{(u,v) \in \mathcal{D}} \log p\left(v \mid \mathcal{H}^u, x^u\right)
\end{equation}

However, directly modeling the distribution $p\left(v \mid \mathcal{H}^u, x^u\right)$ is challenging due to the significant semantic gap between the linguistic representation from dialogue $x^u$ and the behavior patterns in $\mathcal{H}^u$. Standard approaches that simply concatenate these heterogeneous modalities often fail to capture the intrinsic dependencies between them. To resolve this, we posit that both linguistic expressions and behavioral patterns are surface manifestations grounded in a shared latent user intent, denoted as a latent variable $z$. By introducing $z$ as the hidden bridge between these two modalities, we can reformulate the predictive distribution in Equation \ref{crs_raw} via marginalization over the latent space:
\begin{equation}\label{conv_y}
p(v \mid \mathcal{H}^u, x^u) = \int_z p(v \mid z, \mathcal{H}^u, x^u) \cdot p(z \mid \mathcal{H}^u, x^u)\, dz
\end{equation}
Here, $\mathcal{H}^u, x^u, v$ represent the historical interaction sequence, dialogue context and the target item, respectively.

\subsubsection{\textbf{Latent Intent Representation}}\label{intentrepresentation}
Directly optimizing the integral in Equation \ref{conv_y} is intractable due to the infinite nature of the continuous latent space.
To address this, we utilize vector quantization which discretizes the latent space by introducing a set of anchor vectors that serve as a representative basis. We define $z$ as a categorical variable over a set of discrete intent prototypes, denoted as $\mathcal{M} = \{m_k\}_{k=1}^K$. This allows us to approximate the intractable integral as a tractable probability distribution function over these behaviorally grounded anchors:
\begin{equation}\label{conv_y_sum}
p(v \mid \mathcal{H}^u, x^u) \approx \sum_{m_k \in \mathcal{M}} p(v \mid m_k, \mathcal{H}^u, x^u) \cdot p(m_k \mid \mathcal{H}^u, x^u)
\end{equation}
This formulation transforms the intractable abstract intent inference into a tractable probability estimation task by estimating the distribution over these anchor vectors.
To ensure these anchor vectors are meaningful, we implement this quantization via K-Means clustering. We construct a global pool of behavioral contexts $\mathcal{H} = \{ \mathbf{h}_{\tau} \mid \tau \in \mathcal{H}^u, \forall u \in \mathcal{U} \}$ by aggregating different behavior sequence embeddings from all sessions across users. We then apply K-Means clustering to identify $K$ principal centroids, which serve as anchor vectors to represent the latent user space. The K-Means intent centroids (or anchors) are not fixed after their initial offline setup. During the M-step of joint training, these anchors are dynamically updated using gradient descent. This approach ensures that the discrete latent space can adapt to changes in semantics, which are introduced by the semantic embeddings of the LLM.

\subsubsection{\textbf{Variational Inference for User Intent}}
With the anchor set $\mathcal{M}$ established as the discrete basis of our latent space, our objective is to maximize the log-likelihood of the target item $v$. While the reformulation in Equation \ref{conv_y_sum} replaces the infinite integral with a finite summation, direct optimization remains problematic. Specifically, calculating the marginal likelihood requires the true posterior $p(m_k \mid v, \mathcal{H}^u, x^u)$, which is computationally prohibitive to evaluate as it depends on the distribution of all possible items in the catalog.
To address this, we adopt variational inference and introduce a variational distribution $q(m_k \mid \mathcal{H}^u)$ to approximate the true posterior. Instead of optimizing the likelihood directly, we maximize its lower bound, known as the Evidence Lower Bound (ELBO). A proof is provided in Appendix \ref{abb:variations_prove}:
\begin{equation}\label{elbo}
\begin{aligned}
    \text{ELBO} = \sum_{m_j \in \mathcal{M}} q&\bigl(m_j \mid \mathcal{H}^u\bigr) \Bigl(
    \log p\bigl(v \mid m_j, \mathcal{H}^u, x^u\bigr) \\
    &+ \log p\bigl(m_j \mid \mathcal{H}^u, x^u\bigr)
    - \log q\bigl(m_j \mid \mathcal{H}^u\bigr)\Bigr)
\end{aligned}
\end{equation}
This formulation effectively disentangles the optimization into two tasks: estimating the user intent distribution ($q$) and generating recommendations based on that intent distribution ($p$). In the following section, we detail how to parameterize these distributions.

\subsection{\textbf{Intent Inference Step}}
The goal of the Intent Inference Step (E-step) is to refine the inference model $q_\phi(m \mid \mathcal{H}^u)$, parameterized by $\phi$, to accurately estimate the user's latent intent distribution. We first detail the inference model architecture, followed by the optimization objectives.

\subsubsection{\textbf{Inference Model}}
To capture user intent from observable context, we employ a pre-trained sequential recommendation model as the backbone. Let $\mathcal{H}^u$ denote a specific behavior sequence. We first process this sequence through the recommendation model to obtain a dense user behavior representation $\mathbf{h}^u$.
To map this representation to the user intent space, we compute the probability distribution over the set of intent anchors $\mathcal{M}$. The intent distribution estimated by the inference model is computed as:
\begin{equation}\label{q_mu}
q_\phi(m \mid \mathcal{H}^u) = \text{Softmax}\left( \frac{\mathbf{h}^u \mathbf{W}_q \mathbf{M}^\top}{\sqrt{d}} \right)
\end{equation}
where $\mathbf{W}_q$ is learnable parameters, $\mathbf{M} \in \mathbb{R}^{K \times d}$ denotes the dense representations of $K$ distinct intent anchors, each of dimension $d$.

\subsubsection{\textbf{Intent Inference Step Optimization}}
Our objective is to refine the inference parameters $\phi$ to accurately estimate the user's intent. In this step, we fix the parameters $\theta$ of the generative model and exclusively optimize the inference model parameters $\phi$.
The primary objective is to maximize the ELBO. This is equivalent to minimizing the KL divergence between the estimated intent distribution $q_\phi(m \mid \mathcal{H}^u)$ and the posterior distribution implied by the fixed generative model, denoted as $\bar{p}_\theta(m \mid v, \mathcal{H}^u, x^u)$. Conceptually, the generative model acts as a ``teacher'' utilizing the target item $v$ to reveal the actual user intent.
\begin{equation}
    \bar{p}_\theta(m_k \mid v, \mathcal{H}^u, x^u) = \frac{\exp\left(s(v, m_k, \mathcal{H}^u, x^u)\right) \cdot \bar{p}_\theta(m_k \mid \mathcal{H}^u, x^u)}{\sum\limits_{m_j \in \mathcal{M}} \exp\left(s(v, m_j, \mathcal{H}^u, x^u)\right) \cdot \bar{p}_\theta(m_j \mid \mathcal{H}^u, x^u)}
\end{equation}
\begin{equation}
    \mathcal{L}_E = \sum_{(u,v) \in \mathcal{D}} \text{KL}\Bigl(q_\phi(m \mid \mathcal{H}^u) \,\big\|\, \bar{p}_\theta(m \mid v, \mathcal{H}^u, x^u)\Bigr)
\end{equation}
where $\mathcal{D}$ represents the dataset of user-item pairs $(u, v)$. $\bar{p}_\theta(m_k \mid \mathcal{H}^u, x^u)$ and $s(v, m_k, \mathcal{H}^u, x^u)$ correspond to Equations \ref{p_prior} and \ref{score_func}, respectively. Here, we interpret $s(v, m_k, \mathcal{H}^u, x^u)$ as proportional to the likelihood $\bar{p}_\theta(v \mid m_k, \mathcal{H}^u, x^u)$, which is detailed in Section \ref{mstep}.

\subsection{\textbf{Recommendation Learning Step}} \label{mstep}
In the Recommendation Learning Step (M-step), we freeze the inference parameters $\phi$ and optimize the generative parameters $\theta$. Our objective is to maximize the expected log-likelihood of the observed interactions with respect to the variational distribution fixed in the E-step. Referring back to the ELBO formulation (Equation \ref{elbo}), this optimization entails learning two distinct distributions: the intent prior $p_\theta(m \mid \mathcal{H}^u, x^u)$, which is trained to predict the latent intent distribution inferred by the E-step and the generative likelihood $p_\theta(v \mid m, \mathcal{H}^u, x^u)$, which is trained to rank the ground-truth item $v$ highly conditioned on the specific intent $m$.

\subsubsection{\textbf{Generative Model Architecture}}
As shown in the right part of Figure \ref{fig:structure}, the generative model consists of two components: the Intent Prior Module to estimate $p_\theta(m \mid \mathcal{H}^u, x^u)$ and the Item Scoring Module to estimate $p_\theta(v \mid m, \mathcal{H}^u, x^u)$. 
For the Intent Prior Module, we fuse the behavior representation $\mathbf{h}^u$ with the semantic dialogue embedding $\mathbf{x}^u$. We extract $\mathbf{x}^u$ using an LLM-based encoder following recent approaches~\cite{jiang-etal-2024-scaling}.
\begin{equation}\label{p_prior}
p_\theta(m_k \mid \mathcal{H}^u, x^u) = \text{Softmax}\left( \text{FFN}([\mathbf{h}^u; \mathbf{x}^u]) \mathbf{W}_m \mathbf{m}_k^\top \right)
\end{equation}
where $[\cdot;\cdot]$ denotes concatenation, $\mathbf{W}_m$ is a learnable projection matrix, and $\mathbf{m}_k$ is the embedding of the $k$-th intent anchor.

For the Item Scoring Module, while our probabilistic objective requires estimating $p_\theta(v \mid m_k, \mathcal{H}^u, x^u)$, directly computing this distribution is computationally intractable because it requires normalization over the entire item space $\mathcal{V}$. Moreover, in recommendation scenarios, determining the correct ranking order of items is often more critical than obtaining absolute probabilities. Consequently, instead of modeling the normalized distribution, we define an unnormalized score function $s(v, m_k, \mathcal{H}^u, x^u)$.
\begin{equation}\label{score_func}
s(v, m_k, \mathcal{H}^u, x^u) = \left( \mathbf{W}_v \mathbf{v} \right)^\top \left( \text{FFN}([\mathbf{h}^u; \mathbf{x}^u]) \odot (\mathbf{W}_z \mathbf{m}_k) \right)
\end{equation}
Here, $\mathbf{v}$ is the item embedding, and $\odot$ denotes element-wise multiplication, fusing the context with the specific intent prototype.

\subsubsection{\textbf{Recommendation Learning Optimization}}
The goal of the recommendation learning step is still to maximize the ELBO which requires estimating $p_\theta(m \mid \mathcal{H}^u, x^u)$, and $p_\theta(v \mid m, \mathcal{H}^u, x^u)$. As shown above, directly estimating the distribution $p_\theta(v \mid m_k, \mathcal{H}^u, x^u)$ is computationally intractable. Therefore, drawing inspiration from InfoNCE \cite{DBLP:journals/corr/abs-1807-03748}, we optimize with score function $s(v, m_k, \mathcal{H}^u, x^u)$ via negative sampling. For a given intent $m_k$ (sampled from $q_\phi$), the model must distinguish the ground-truth item $v$ from a set of negative items $\mathcal{V}_{neg}$ sampled uniformly and treated as the noise distribution. The ELBO optimization can be transformed into optimizing the loss $\mathcal{L}_M$. Proofs are shown in Appendix \ref{abb:infonce_prove}:
\begin{equation}\label{loss_infonce}
\mathcal{L}_{NCE} = - \sum_{(u,v)} \sum_{m_k} \bar{q}_\phi(m_k \mid \mathcal{H}^u) \log \frac{\exp(s(v, m_k, \mathcal{H}^u, x^u))}{\sum\limits_{v' \in \{v\} \cup \mathcal{V}_{neg}} \exp(s(v', m_k, \mathcal{H}^u, x^u))}
\end{equation}
\begin{equation}\label{loss_M}
    \mathcal{L}_M = \mathcal{L}_{NCE} + \lambda \, \text{KL}\bigl(\bar{q}_\phi(m \mid \mathcal{H}^u) \,\|\, p_\theta(m \mid \mathcal{H}^u, x^u)\bigr)
\end{equation}
Here, $(u,v) \in \mathcal{D}$ represents the user-item pair in the dataset $\mathcal{D}$, $m_k \in \mathcal{M}$ represents the $k$-th user intent anchor. $\bar{q}_\phi(m_k \mid \mathcal{H}^u)$ represents the fixed inference model shown in Equation \ref{q_mu}. $p_\theta(m_j \mid \mathcal{H}^u, x^u)$ and $s(v, m_k, \mathcal{H}^u, x^u)$ are shown in Equation \ref{p_prior} and \ref{score_func}, respectively. We introduce the weight parameter $\lambda$ to control the balance between the Item Scoring Module and Intent Prior Module. This differs slightly from the standard ELBO, but we find empirically that it improves training efficiency and performance.

\subsection{Optimization and Inference of \ours}
Here, we propose strategies designed to ensure robust optimization and strong performance. We propose a warm-up procedure to ensure training stability. For inference, we introduce post-hoc semantic filtering and reranking strategies that use LLMs' semantic reasoning ability for better recommendation results. 

\subsubsection{\textbf{Warm-up Strategy}}\label{warmup}
Initiating the Variational EM process with randomly initialized parameters often leads to inefficient optimization, as the unaligned inference and generative modules provide noisy supervision to one another. To mitigate this, we employ a warm-up strategy to pre-train both components before joint optimization.\\
\noindent \textbf{Inference Model Warm-up.} Although the actual user intent is unobservable, accurate user intent estimation should result in accurate recommendations. Therefore, we propose an assistant recommendation module with parameters $\psi$ to make recommendations based on the estimated user intent distribution, represented as $q_\psi(v | m)$. Details are shown in Appendix \ref{abb_assrec}. Similar to the negative sampling strategy in Recommendation Learning Step, we calculate the assistant loss as:
\begin{equation}
    \hat{q}(v | \mathcal{H}^u) = \sum_{m \in \mathcal{M}} q_\psi(v | m) q_\phi(m \mid \mathcal{H}^u)
\end{equation}
\begin{equation}\label{eq:rec_e}
\mathcal{L}^{rec}_E = - \sum_{(u,v) \in \mathcal{D}} \log \frac{\exp(\hat{q}(v \mid \mathcal{H}^u))}{\sum\limits_{v' \in \{v\} \cup \mathcal{V}_{neg}} \exp(\hat{q}(v' \mid \mathcal{H}^u))}
\end{equation}
\noindent \textbf{Generative Model Warm-up.} For the generative model, the warm-up process serves a dual purpose: it ensures the model is competent at the recommendation task (Item Retrieval) while simultaneously aligning the Intent Prior with the intent estimation from the inference model. Unlike the inference stage, the generative model does not require an external assistant module. It can perform standalone recommendation by marginalizing over its own prior distribution.
\begin{equation}
\hat{s}_\theta(v \mid \mathcal{H}^u, x^u) = \sum_{m_k \in \mathcal{M}} s(v, m_k, \mathcal{H}^u, x^u) \cdot {p_\theta(m_k \mid \mathcal{H}^u, x^u)}
\end{equation}
\begin{equation} 
\mathcal{L}^{rec}_{M} = - \sum_{(u,v) \in \mathcal{D}} \log \frac{\exp(\hat{s}_\theta(v \mid \mathcal{H}^u, x^u))}{\sum\limits_{v' \in \{v\} \cup \mathcal{V}_{neg}} \exp(\hat{s}_\theta(v' \mid \mathcal{H}^u, x^u))} 
\end{equation}
\begin{equation}\label{eq:ass_m}
    \mathcal{L}^{ass}_{M} = \mathcal{L}_M + \alpha_M \mathcal{L}^{rec}_{M}
\end{equation}
where $p_\theta(m_k \mid \mathcal{H}^u, x^u)$ and $s(v, m_k, \mathcal{H}^u, x^u)$ are defined in Equation \ref{p_prior} and \ref{score_func}, respectively. The loss function $\mathcal{L}_M$ is shown in Equation \ref{loss_M}. The hyperparameter $\alpha_M$ balances the objectives of the assistant recommendation task and ELBO.

\begin{algorithm}[t]
\small
  \SetKwInOut{Input}{Input}
  \SetKwInOut{Output}{Output}
  \KwIn{User set $\mathcal{U}$, item set $\mathcal{V}$, behavior history $\mathcal{H}^u$, dialogue context $x^u$.}
  \KwOut{Recommendation scores $\hat{s}_\theta(v \mid \mathcal{H}^u, x^u)$ for target items.}

  \tcp{Warm-up Strategy}
  \While{Inference Warm-up not converged}{
      Update inference parameters $\phi$ and assistant parameters $\psi$ using $\mathcal{L}^{rec}_E$ (Eq. \ref{eq:rec_e}).
  }
  \While{Generative Warm-up not converged}{
      Update generative parameters $\theta$ using $\mathcal{L}^{ass}_{M}$ (Eq. \ref{eq:ass_m}).
  }

  \tcp{Variational Optimization}
  \While{Joint Training not converged}{
      \tcp{\textbf{E-step: Intent Inference}}
      Compute target posterior $ \bar{p}_\theta(m \mid v, \mathcal{H}^u, x^u)$ using current generative model $\theta$\;
      Update inference parameters $\phi$ to minimize $\mathcal{L}_E^{total} = \mathcal{L}_{E} + \alpha_E \mathcal{L}^{rec}_E$\;

      \tcp{\textbf{M-step: Recommendation Learning}}
      Sample latent intents from $q_\phi(m_k \mid \mathcal{H}^u)$ using current inference model $\phi$\;
      Update generative parameters $\theta$ to minimize $\mathcal{L}^{total}_M = \mathcal{L}_{M} + \alpha_M \mathcal{L}^{rec}_M$\;
  }
  \Return Recommendation scores $\hat{s}_\theta(v \mid \mathcal{H}^u, x^u)$, which induce the predictive distribution $p_\theta(v \mid \mathcal{H}^u, x^u)$ after normalization.
  \caption{Optimization Procedure of \ours}
  \label{algo1}
\end{algorithm}

\subsubsection{\textbf{Optimization of \ours}} \label{optimization_sec}
Algorithm \ref{algo1} outlines the complete training procedure of \ours. In the warm-up phase, both inference model and generative model are pretrained for their respective tasks. Once initialized, we proceed to the joint Variational EM phase, where the two modules iteratively update to maximize the Evidence Lower Bound (ELBO). In inference learning step (E-step), we fix the generative parameters $\theta$ and update the inference parameters $\phi$. Crucially, we retain the assistant recommendation loss $\mathcal{L}^{rec}_E$ in the total objective $\mathcal{L}_E^{total} = \mathcal{L}_{E} + \alpha_E \mathcal{L}^{rec}_E$. The auxiliary term functions as a discriminative regularizer, enforcing the principle that the inferred intent must be sufficient to explain the user's item choice, independent of the generative prior. By compelling the aggregated intent distribution $\hat{q}(v|\mathcal{H}^u)$ to accurately predict the target item $v$, we impose a constraint that prevents the inference model from degenerating into a uniform or uninformative distribution, thereby mitigating posterior collapse.
In the recommendation learning step (M-step), we fix the inference parameters $\phi$ and update the generative parameters $\theta$. Similarly, we also incorporate the auxiliary target into the M-step objective $\mathcal{L}^{total}_M = \mathcal{L}_{M} + \alpha_M \mathcal{L}^{rec}_M$. This alternating procedure continues until convergence, resulting in a framework where the latent user intent serves as an effective bridge between the behavioral modeling and the semantic understanding ability of the LLM.

\subsubsection{\textbf{Inference Procedure of \ours}}
During the inference phase, we propose three progressive variants to systematically leverage LLM capabilities: 
\textbf{Base Inference} ($\text{\ours}_B$): the LLM is used only as the semantic normalizer and encoder without additional reasoning steps.
\textbf{Filtered Inference} ($\text{\ours}_F$): we enhance \ours by introducing a Hard-Filter mechanism to account for explicit user constraints (e.g., ``I prefer animated movies''). Here, the LLM is prompted to extract mandatory attributes (e.g., genre, artist, category) from the dialogue, and the recommended item list is pruned to exclude items that violate these constraints.
\textbf{Verified Inference} ($\text{\ours}_V$): the LLM is used as a ``critic'' to evaluate the consistency between the user's request and the recommended items, and then rerank the item list. These variants ($\text{\ours}_B \rightarrow \text{\ours}_F \rightarrow \text{\ours}_V$) represent different strategies for incorporating LLM-driven parsing, constraint enforcement, and semantic verification in the inference process.

\begin{table*}[htbp]
  \centering
  \caption{Performance on full datasets in the single-turn setting. Due to high computational costs, certain baselines are limited to the sampled dataset (see Table \ref{tab:one_turn_llm}). \textbf{Bold} and \underline{underlined} values denote the best and second-best results, respectively. Asterisks (*) indicate statistically significant improvements of \ours over the baselines under a t-test with $p < 0.05$.}
\resizebox{\linewidth}{!}{
\begin{tabular}{l|cccc|cccc|cccc}
\toprule
      & \multicolumn{4}{c|}{MovieLens-1M} & \multicolumn{4}{c|}{VideoGames} & \multicolumn{4}{c}{CDs} \\
      & Recall@5 & NDCG@5 & Recall@20 & NDCG@20 & Recall@5 & NDCG@5 & Recall@20 & NDCG@20 & Recall@5 & NDCG@5 & Recall@20 & NDCG@20 \\
\midrule
CRM   & 0.1165 & 0.0690 & \underline{0.3273} & \underline{0.1279} & 0.0024 & 0.0014 & 0.0090 & 0.0032 &  0.0089     &    0.0055   &      0.0297  & 0.0111 \\
UNICORN & \underline{0.1215} & \underline{0.0734} & 0.3086 & 0.1274 & \underline{0.1138} & \underline{0.0753} & \underline{0.2317} & \underline{0.1099} & \textbf{0.0839} & \textbf{0.0518} & \textbf{0.1911} & \textbf{0.0833} \\
CRIF  & 0.0784 & 0.0489 & 0.2363 & 0.0926 & 0.0427 & 0.0267 & 0.1058 & 0.0443 & 0.0157 & 0.0100 & 0.0442 & 0.0180 \\
\midrule
$\text{\ours}_B$ & \textbf{0.1437*} & \textbf{0.0921*} & \textbf{0.3305*} & \textbf{0.1449*} & \textbf{0.1378*} & \textbf{0.0964*} & \textbf{0.2598*} & \textbf{0.1310*} & \underline{0.0485} & \underline{0.0326} & \underline{0.1083} & \underline{0.0494} \\
\bottomrule
\end{tabular}
}
\label{tab:one_turn_traditional}%
\end{table*}

\begin{table*}[htbp]
  \centering
  \caption{Performance on sampled datasets in the single-turn setting. \textbf{Bold} and \underline{underlined} values denote the best and second-best, respectively. Asterisks (*) indicate statistically significant improvements of \ours over the baselines under a t-test with $p < 0.05$.}
  \resizebox{\linewidth}{!}{
    \begin{tabular}{l|cccc|cccc|cccc}
    \toprule
          & \multicolumn{4}{c|}{$\text{MovieLens}_{sample}$} & \multicolumn{4}{c|}{$\text{VideoGames}_{sample}$} & \multicolumn{4}{c}{$\text{CDs}_{sample}$} \\
          & Recall@5 & NDCG@5 & Recall@20 & NDCG@20 & Recall@5 & NDCG@5 & Recall@20 & NDCG@20 & Recall@5 & NDCG@5 & Recall@20 & NDCG@20 \\
    \midrule
    Llama-3.1-8B & {0.054} & {0.0396} & {0.092} & 0.0507 & {0.020} & {0.0143} & {0.034} & 0.0182 & {0.012} & {0.0074} & {0.016} & {0.0085} \\
   Gemini-1.5-Pro & 0.068 & 0.0414 & 0.096 & 0.0604 & 0.034 & 0.0216 & 0.062 & 0.0409 & 0.018 & 0.0136 & 0.088 & 0.0433 \\
    GPT-4  & 0.064 & 0.0421 & 0.102 & 0.0622 & 0.032 & 0.0204 & 0.058 & 0.0400 & 0.028 & 0.0204 & 0.104 & 0.0476 \\
   Chat-Rec & {0.086} & {0.0546} & {0.150} & 0.0950 & {0.038} & {0.0268} & {0.080} & 0.0630 & {0.016} & {0.0111} & {\underline{0.108}} & {\textbf{0.0787}} \\
   InteRecAgent & {0.006} & {0.0043} & {0.022} & 0.0087 & {0.014} & {0.0086} & {0.024} & 0.0118 & {0.018} & {0.0129} & {0.030} & {0.0163} \\
   UniCRS & 0.092 & 0.0551 & 0.277 & 0.1067 & 0.078 & 0.0594 & 0.223 & 0.1291 & 0.038 & 0.0294 & 0.082 & 0.0389 \\
   CRAG & 0.104 & 0.0823 & 0.220 & 0.0849 & 0.040 & 0.0296 & 0.188 & 0.1011 & 0.034 & 0.0249 & 0.104 & 0.0469 \\
    \midrule
    $\text{\ours}_B$ & 0.142* & 0.0879* & \textbf{0.340*} & 0.1426* & 0.140* & 0.1008* & \textbf{0.250*} & 0.1326* & 0.040 & 0.0270 & 0.106 & 0.0454 \\
    $\text{\ours}_F$ & \textbf{0.180*} & \textbf{0.1209*} & 0.318* & \textbf{0.1604*} & \underline{0.146*} & {\underline{0.1072*}} & {0.230} & \underline{0.1347*} & \textbf{0.058*} & \textbf{0.0375*} & \textbf{0.126*} & \underline{0.0573} \\
    $\text{\ours}_V$ & \underline{0.160*} & \underline{0.1139*} & \underline{0.320*} & \underline{0.1582*} & \textbf{0.180*} & \textbf{0.1386*} & \underline{0.246*} & \textbf{0.1592*} & \underline{0.048*} & \underline{0.0302*} & \underline{0.108} & 0.0489 \\
    \bottomrule
    \end{tabular}}
  \label{tab:one_turn_llm}%
\end{table*}

\section{Experiments}
In this section, we conduct extensive experiments to address the following research questions:\\
\textbf{Q1:} How does \ours perform for conversational recommendation by bridging the gap between LLM-based semantic reasoning and user behavior modeling?\\
\textbf{Q2:} How do the core components and hyperparameters affect \ours?\\
\textbf{Q3:} Does the latent intent effectively serve as the bridge? Are the learned intents interpretable?\\
\textbf{Q4:} How does \ours compare to other LLM-based approaches in terms of efficiency?

\begin{table}[htbp]
  \centering
  \caption{Performance on sampled datasets under multi-turn settings. \textbf{Bold} and \underline{underlined} values denote the best and second-best results, respectively. Asterisks (*) indicate statistically significant improvements of \ours over the baselines under a t-test with $p < 0.05$.}
  \resizebox{\linewidth}{!}{
    \begin{tabular}{l|ccc|ccc|ccc}
    \toprule
          & \multicolumn{3}{c|}{$\text{MovieLens}_{sample}$} & \multicolumn{3}{c|}{$\text{VideoGames}_{sample}$} & \multicolumn{3}{c}{$\text{CDs}_{sample}$} \\
          & S@3   & S@5   & AT    & S@3   & S@5   & AT    & S@3   & S@5   & AT \\
    \midrule
    CRM   &    0.413   &   -    &    -   &   0.013    &   -    &    -   &    0.042   &   -   &  - \\
    UNICORN & 0.248 & 0.378 & 4.174 & 0.184 & 0.260 & 4.362 & 0.174 & 0.232 & 4.406 \\
    CRIF  & 0.266 & 0.448 & 4.064 & 0.126 & 0.216 & 4.578 & 0.130 & 0.208 & 4.610 \\
    \midrule
    Llama-3.1-8B & {0.188} & {0.540} & 3.624 & {0.208} & {0.276} & 4.668 & {0.000} & {0.258} & 4.832 \\
    Gemini-1.5-Pro & 0.224 & 0.580 & 3.484 & 0.248 & 0.384 & 4.108 & 0.170 & 0.498 & 4.412 \\
    GPT-4  & 0.246 & 0.602 & 3.472 & 0.236 & 0.398 & 4.246 & \underline{0.184} & \underline{0.548} & 4.254 \\
    Chat-Rec & {0.212} & \textbf{0.686} & 3.618 & 0.224 & \textbf{0.724} & \underline{3.760} & 0.146 & \textbf{0.558} & \textbf{4.202} \\
    InteRecAgent & 0.070 & 0.230 & 4.754 & 0.156 & 0.292 & 4.492 &    0.102   &   0.230    & 4.754 \\
    UniCRS & 0.376 & 0.622 & 3.578 & 0.272 & 0.394 & 3.952 & 0.164 & 0.324 & 4.590 \\
    CRAG & 0.348 & 0.582 & 3.874 & 0.262 & 0.422 & 4.510 & 0.056 & 0.364 & 4.752 \\
    \midrule
    $\text{\ours}_B$ & 0.388 & 0.482 & 3.760 & 0.264 & 0.336 & 4.078 & 0.100 & 0.142 & 4.642 \\
    $\text{\ours}_F$ & \underline{0.472*} & 0.562 & \underline{3.424*} & \underline{0.286*} & 0.356 & 4.042 & \textbf{0.190*} & 0.258 & \underline{4.236} \\
    $\text{\ours}_V$ & \textbf{0.532*} & \underline{0.650} & \textbf{3.298*} & \textbf{0.378*} & \underline{0.426} & \textbf{3.706*} & 0.182 & 0.220 & 4.422 \\
    \bottomrule
    \end{tabular}}
  \label{tab:multiturn}
\end{table}%

\subsection{Experimental Settings}
Evaluating Conversational Recommender Systems (CRS) is challenging due to the high cost of real-world user studies. Following established protocols \cite{zhang2024usimagent, yoon2024evaluating}, we leverage Gemini-1.5-Pro \cite{team2024gemini} to simulate diverse user behaviors via role-playing prompts (detailed in Appendix \ref{abb:prompts}). To ensure a rigorous and fair comparison, all baselines are evaluated with the same user simulator. Our evaluation spans two distinct settings: Single-Turn Recommendation, where the system must respond to a standalone natural language query, and Multi-Turn Recommendation, which allows for iterative refinement until satisfactory items are found \cite{jin2023lending, huang2023recommender}.
To address possible concerns about the bias of Gemini-1.5-Pro as a user simulator, we further conduct a blind human evaluation. We collect 30 multi-turn sessions from human users, with 10 sessions each for LatentCRS, InteRecAgent, and Chat-Rec, and compare them with 30 simulator-generated sessions under the same conditions. Two independent annotators, who are blinded to the source of each dialogue, assess which dialogue is more satisfactory. The annotators preferred the simulator-generated dialogues in 43\% and 47\% of the cases, respectively. This near-balanced preference indicates that the simulator-generated dialogues are close to human-generated dialogues in terms of perceived satisfaction.

\noindent \textbf{Datasets.}
We conduct experiments on three distinct real-world datasets \cite{hou2024bridging}: MovieLens-1M (MovieLens)\footnote{https://grouplens.org/datasets/movielens/1m/}, Amazon VideoGames (VideoGames), and Amazon CDs and Vinyl (CDs) \cite{hou2024bridging}. To manage computational costs for LLM-based methods and multi-turn simulations, we follow prior work \cite{huang2023recommender} by sampling 500 test instances per dataset, denoted as $\text{MovieLens}_{sample}$, $\text{VideoGames}_{sample}$ and $\text{CDs}_{sample}$. More details are shown in Appendix \ref{abb:dataset}.

\noindent \textbf{Baselines.}
To evaluate the effectiveness of \ours, we compare it with two categories of methods.\\
Traditional Conversational Recommendation Models: 
\textbf{CRM} \cite{SunZ18} feeds the belief tracker results to an FM-based recommendation method to integrate the conversational and recommender components. 
\textbf{UNICORN} \cite{deng2021unified}  formulates the conversational recommendation problem as a unified policy learning task.
\textbf{CRIF} \cite{hu2022learning} utilizes a four-phase process consisting of offline representation learning, tracking, decision-making, and inference.\\
LLM-based Conversational Recommendation Methods:
\textbf{Llama-3.1-8B} \cite{dubey2024llama}, \textbf{Gemini-1.5-Pro} \cite{team2024gemini}, \textbf{GPT-4} \cite{achiam2023gpt}: These are representative LLMs, including both open-source and closed-source models.
\textbf{Chat-Rec} \cite{gao2023chat} converts user profiles and historical interactions into prompts to help LLMs build CRS.
\textbf{InteRecAgent} \cite{huang2023recommender} employs LLMs as the core processing unit and recommender models as tools.
\textbf{UniCRS} \cite{wang2022towards} formulates both recommendation and response generation as knowledge-enhanced prompt learning over a fixed pre-trained language model.
\textbf{CRAG} \cite{zhu2025collaborative} is an LLM-based CRS method that augments response generation with collaborative filtering signals. \\
Here, GPT-4, Gemini-1.5-Pro, and Llama-3.1-8B were evaluated in a few-shot prompt setting without fine-tuning. Chat-Rec and InteRecAgent were explicitly designed to be knowledge-grounded in our experiments. These two baselines were equipped with the same recommender backbone (ICLRec) used by LatentCRS, ensuring they had access to global collaborative knowledge. Both LatentCRS and UniCRS used the same Llama-3.1-8B backbone for fair comparison.

\noindent \textbf{Metrics.}
For single-turn conversational recommendation, the objective is to rank candidate items based on user interests. Therefore, we focus on the widely used Recall@K and Normalized Discounted Cumulative Gain (NDCG)@K ranking metrics.
In the multi-turn conversational recommendation setting, the goal is to identify the items the user desires within as few conversation turns as possible. Following \cite{gao2023chat, huang2023recommender}, we use the Average Success Rate (S@K) and Average Turns (AT). S@K represents the success rate of recommending the target item within K turns, while AT indicates the average number of turns to achieve successful recommendations.

\noindent \textbf{Implementation Details.}
Within our framework, we use Llama-3.1-8B-Instruct \cite{dubey2024llama} as the incorporated LLM and ICLRec \cite{chen2022intent} as the sequential recommendation model. More details are shown in Appendix \ref{app_impledetails}, and the code is available on GitHub\footnote{\url{https://github.com/kylokano/LatentCRS}}.

\begin{table*}[t]
  \centering
  \caption{Ablation study. The results are reported on the full datasets with basic inference strategy, which generates recommendations directly from the generative model without post-hoc LLM reranking or filtering.}
  \resizebox{\linewidth}{!}{
    \begin{tabular}{l|cccc|cccc|cccc}
    \toprule
          & \multicolumn{4}{c|}{MovieLens-1M} & \multicolumn{4}{c|}{VideoGames} & \multicolumn{4}{c}{CDs} \\
          & Recall@5 & NDCG@5 & Recall@20 & NDCG@20 & Recall@5 & NDCG@5 & Recall@20 & NDCG@20 & Recall@5 & NDCG@5 & Recall@20 & NDCG@20 \\
    \midrule
    $\text{\ours}_B$ & 0.1437 & 0.0921 & 0.3305 & 0.1449 & 0.1378 & 0.0964 & 0.2598 & 0.1310 & 0.0485 & 0.0326 & 0.1083 & 0.0494 \\
    \midrule
    w/o Inference Module & 0.0232 & 0.0147 & 0.0678 & 0.0271 & 0.0309 & 0.0204 & 0.0595 & 0.0283 & 0.0240 & 0.0150 & 0.0667 & 0.0269 \\
    Direct KL Alignment & 0.0847 & 0.0518 & 0.2256 & 0.0911 & 0.0728 & 0.0472 & 0.1740 & 0.0756 & 0.0258 & 0.0165 & 0.0689 & 0.0285 \\
    w/o RecLoss & 0.0755 & 0.0698 & 0.2073 & 0.0805 & 0.1006 & 0.0700 & 0.2092 & 0.1098 & 0.0241 & 0.0132 & 0.0650 & 0.0264 \\
    \bottomrule
    \end{tabular}}
  \label{tab:Ablation_Study}%
\end{table*}

\begin{table}[htbp]
  \centering
  \caption{Efficiency of different methods. Inference time is measured only for locally deployed LLMs. For commercial black-box models, we report token counts without measuring runtime. Additional results for the VideoGames and CDs datasets are given in Table \ref{app:cost} in the Appendix.}
  \resizebox{0.75\linewidth}{!}{
    \begin{tabular}{l|ccc}
    \toprule
          & \multicolumn{3}{c}{MovieLens-1M} \\
          & Input (Tokens) & Output (Tokens) & Time (ms)\\
    \midrule
    Llama-3.1-8B &   119.51    &  148.20  &  2568  \\
    Gemini-1.5-Pro &   120.49    &  140.56  &  - \\
    GPT-4  &    118.95   &    152.81   &  - \\
    \midrule
    $\text{\ours}_B$ & 85.50 & 1.00 & 87  \\
    $\text{\ours}_F$ & 317.50 & 35.59 & 709  \\
    $\text{\ours}_V$ & 810.26 & 98.63 & 2168 \\
    \bottomrule
    \end{tabular}}
  \label{tab:cost}%
\end{table}%

\subsection{Overall Performance (RQ1)}
This section evaluates the effectiveness of \ours across three real-world datasets under both single-turn and multi-turn conversational settings.
For the single-turn setting, we evaluate non-LLM-based methods on the full datasets, while LLM-based methods are evaluated on sampled datasets due to their high inference cost. The results are summarized in Table~\ref{tab:one_turn_traditional} and Table~\ref{tab:one_turn_llm}, respectively.
\ours achieves the best or competitive performance on most metrics, especially in ranking accuracy. This result supports our core hypothesis that latent intent modeling can effectively connect collaborative signals with linguistic queries. One exception is that UNICORN achieves strong performance on the Amazon CDs dataset in the single-turn setting, which is mainly due to the specific characteristics of this dataset and the design of UNICORN. Compared with MovieLens-1M, which contains only 18 attributes, Amazon CDs contains 531 attributes. UNICORN relies on explicit attribute querying, which works as a hard deterministic filter over the item space. This mechanism is especially effective for narrow retrieval in the single-turn setting when the attribute space is highly fine-grained. However, this advantage is closely tied to the attribute structure of Amazon CDs and does not generalize well to datasets with fewer attributes or to more natural multi-turn conversations. Moreover, such a setting is less aligned with real-world usage. As shown in our experiments, UNICORN performs much worse on other datasets and in the multi-turn setting. In contrast, direct LLM-based baselines underperform compared with traditional CRS methods. Despite their strong language understanding ability, these models struggle to use historical interaction data when directly prompted. This empirical gap highlights the need to bridge semantic representation and user behavior modeling, which is exactly the problem that \ours aims to address.

For the multi-turn setting, due to the high cost of the user simulator, we compare \ours only on the sampled datasets, with the results shown in Table~\ref{tab:multiturn}. Note that feature-querying baselines such as CRM \cite{SunZ18} are limited by the available attribute categories in the dataset, which restricts their maximum dialogue depth.
\ours shows a clear advantage in interaction efficiency, achieving higher success rates within the first three turns. By jointly using behavioral history and real-time dialogue feedback, our framework rapidly narrows down user intent and reduces the number of interactions needed to identify the target item.

\subsection{Core Components (RQ2)}
In this section, we present a comprehensive analysis of \ours, dissecting the contributions of its core components.

\subsubsection{\textbf{Ablation Study.}}
To evaluate the effectiveness of each component of \ours, we conduct an ablation study on the full datasets with basic inference strategy ($\text{\ours}_B$), as the basic inference isolates the core mechanisms of bridging semantic-behavioral representation without post-hoc LLM filtering or reranking. We compare $\text{\ours}_B$ with the following variations: \textbf{(1) w/o Inference Module}: We remove the inference model entirely and train the generative model directly using only the recommendation objective. \textbf{(2) Direct KL Alignment}: Instead of the alternating EM procedure, we train the inference and generative models simultaneously, enforcing alignment via a standard KL divergence constraint. \textbf{(3) w/o RecLoss}: We retain the EM framework but remove the auxiliary recommendation loss ($\mathcal{L}^{rec}_E, \mathcal{L}^{rec}_M$) from both steps.
The results are presented in Table \ref{tab:Ablation_Study}. The significant drop in the w/o Inference Module variant confirms that the generative model cannot learn a robust latent space from linguistic signals alone. The underperformance of Direct KL Alignment indicates that simultaneous updates lead to optimization instability. This validates that the alternating EM framework is essential for effectively disentangling and aligning the two modalities. The degradation in w/o RecLoss confirms that the auxiliary loss is critical. It functions as a discriminative regularizer, preventing posterior collapse and ensuring the latent space remains behaviorally meaningful.

\subsubsection{\textbf{Inference Strategies.}}
In the single-turn setting, the base strategy ($\text{\ours}_B$) delivers strong independent performance, confirming that the learned latent intent effectively captures user preferences without external reliance. While adequate on its own, further enhancements via semantic filtering ($\text{\ours}_F$) and reranking ($\text{\ours}_V$) yield incremental gains, demonstrating that deeper semantic reasoning serves as a valuable complement to the robust behavioral backbone.
In contrast, the impact of LLM augmentation becomes significantly more pronounced in multi-turn scenarios. Here, $\text{\ours}_F$ consistently outperforms the base model, while the reranking variant $\text{\ours}_V$ typically achieves the highest overall performance. We attribute this to the LLMs' superior capacity for dialogue management—specifically in generating precise clarifying questions and interpreting complex feedback, which creates a positive feedback loop where better system responses elicit more informative user constraints.

\subsubsection{\textbf{Hyperparameter Sensitivity}}
\begin{figure}[t]
  \includegraphics[width=\linewidth]{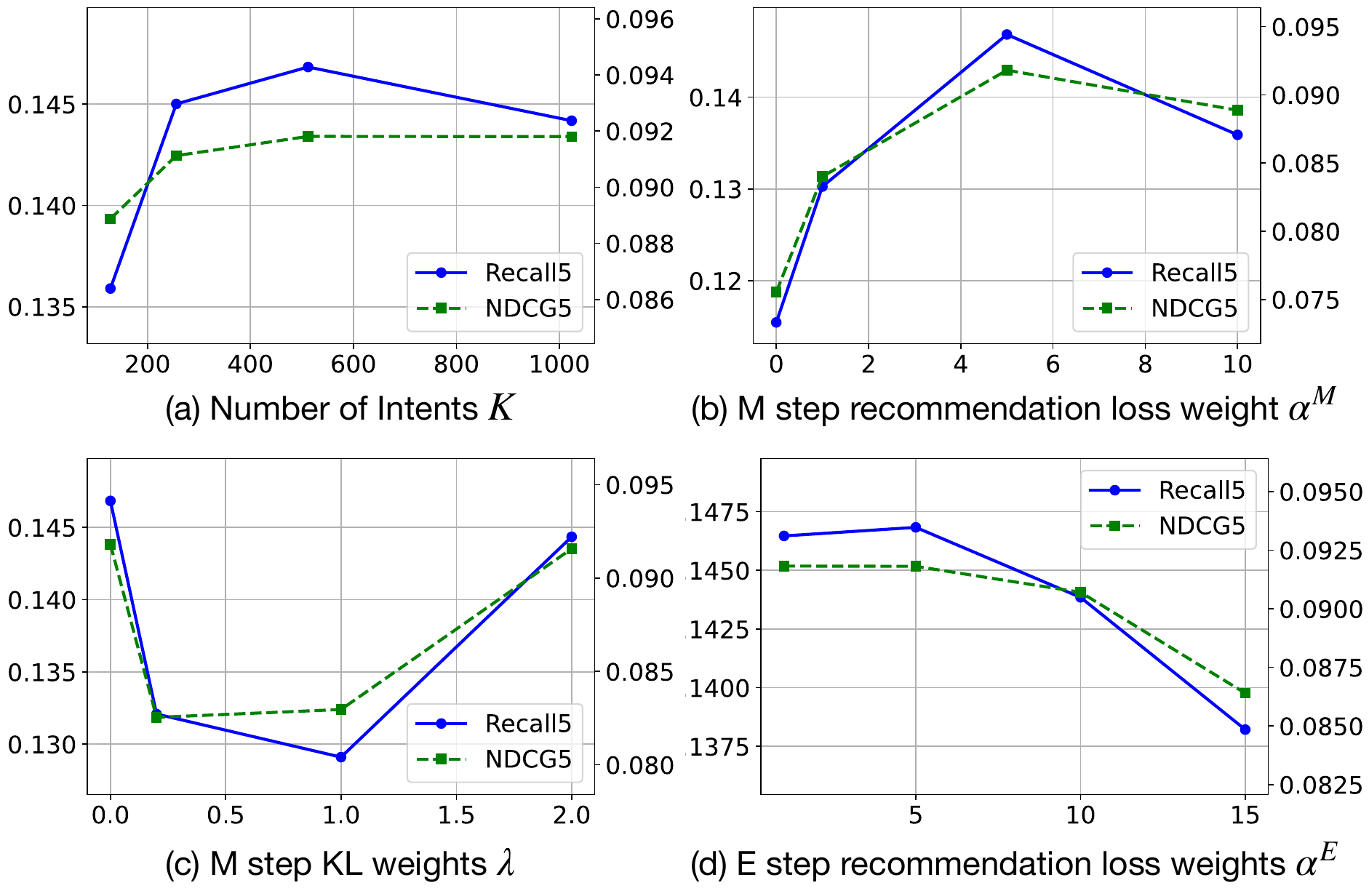}
  \caption{The influence of hyperparameters.}
  \label{fig:hyperparameter}
\end{figure}

In this section, we present experimental results to examine the sensitivity of \ours to key hyperparameters. We evaluate the performance differences with respect to the number of intent anchors $K$ and the loss weights $\alpha_M, \alpha_E$ shown in Section \ref{optimization_sec}. The results are illustrated in Figure \ref{fig:hyperparameter}. Due to space constraints, we only report the results obtained on the MovieLens-1M dataset. From the analysis, we observe that the optimal selection of hyperparameters plays a crucial role in achieving superior performance. For the number of intent anchors $K$ in user intent space, we observe that the model's performance remains highly stable for $K\geq 128$, with metric fluctuations restricted to approximately 1\%. This suggests that \ours is robust to the number of anchors, as long as $K$ reaches a sufficient threshold to capture diverse user behaviors.

\subsection{Visualization of Latent Intent (RQ3)}
In this section, we provide an in-depth analysis of the latent intent mechanism, the core component of \ours. We begin by examining the semantic interpretability of the learned intent anchors, followed by a visualization of how semantic and behavioral representations influence latent user intent.

\subsubsection{\textbf{Meaningful Anchors}}
To validate that the discretized intent prototypes $\mathcal{M}$ capture meaningful and distinct user interests, we analyze the semantic characteristics of the learned anchors. We show representative anchors and display the top-ranked items by conditioning exclusively on each anchor. As illustrated in Table \ref{tab:intent_exp}, the retrieved items for each anchor exhibit strong semantic consistency. This confirms that our framework has successfully disentangled complex user behaviors into coherent, granular intent prototypes. 

\subsubsection{\textbf{Influence of Semantic and Behavioral Representations on Intent}}
\begin{figure}[t]
  \includegraphics[width=\linewidth]{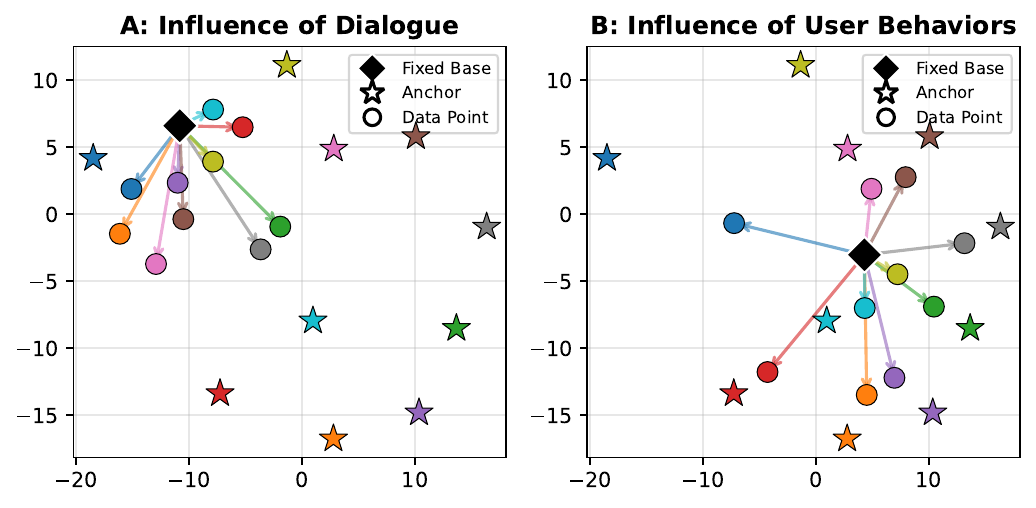}
  \caption{Visualization of Controlled Modulation Analysis. This plot illustrates how varying specific input modalities influences the resulting user intent representation. The \textbf{Black Diamond} (Fixed Base) represents the intent embedding derived from the generic, fixed input. \textbf{Stars} (Anchors) represent the specific semantic or behavioral targets. \textbf{Circles} (Data Points) denote the final intent representations generated by \ours after integrating the varied input. Arrows indicate the shift in the latent space from the generic baseline to the specific representation.}
  \label{fig:behavior_modulation}
\end{figure}

To empirically validate that both semantic and behavioral representations play a vital role in shaping user intent, we conduct a Controlled Modulation Analysis. This method isolates the contribution of each modality by fixing one input while varying the other to observe directional shifts in the latent space.
First, we fix the user behavior to a neutral baseline (represented by a generic history $\mathcal{H}_{generic}$) and pair it with distinct dialogue queries $\{x_k\}$ associated with specific semantic anchors. Conversely, we fix the dialogue query to a generic request ($x_{generic}$) and pair it with diverse behavioral histories $\{\mathcal{H}_k\}$ corresponding to different intent anchors.
As visualized in Figure \ref{fig:behavior_modulation}, we observe that the generated intent vectors diverge significantly from the generic baseline (Fixed Base), shifting systematically toward the latent space regions defined by the variable modality. This trajectory confirms that \ours is not dominated by a single modality; rather, it effectively integrates signals from both semantic and behavioral inputs to produce robust, context-aware intent representations.

\begin{table}[t]
  \centering
  \small
  \fbox{
  \begin{minipage}{0.92\linewidth}
    \centering
    \textbf{Top-ranked Items under Different Intent Anchors}
    
    
    \begin{tabular}{p{0.38\linewidth} p{0.50\linewidth}}
      \toprule
      \textbf{Intent Anchor} & \textbf{Top-ranked Items} \\
      \midrule
      Animated Films for Children 
      & Aladdin; The Little Mermaid; Cinderella; The Land Before Time \\
      \midrule
      Suspense and Violence 
      & Scream; Seven; Cape Fear; I Know What You Did Last Summer \\
      \bottomrule
    \end{tabular}
  \end{minipage}
  }
  \caption{Visualization of intent-space anchors. We display the top-4 items ranked by the generative model when conditioned on specific intent anchors. The examples show that different anchors capture semantically coherent interest clusters, such as animated films for children and suspense/thriller films, validating the interpretability of our discretized latent space.}
  \label{tab:intent_exp}
\end{table}

\subsection{Efficiency Analysis (RQ4)}\label{Efficiency}
Due to the high cost of LLMs, efficiency is a critical factor for LLM-based CRS, especially in practical applications. Since the computational cost of LLMs is significantly higher than that of other components, we also report the total token usage of the model by considering both input and output tokens of the LLMs. The token usage is presented in Table \ref{tab:cost}. We use the Llama 3.1 tokenizer as a proxy for the token counts in GPT-4 and Gemini-1.5-Pro, as we only have access to detokenized text. The following observations can be made:
Existing LLM-based CRS methods often require many LLM input and output tokens, leading to high inference cost.
In contrast, $\text{\ours}_B$ uses LLMs only to obtain the embeddings of user responses and achieves satisfactory performance with fewer tokens. Even $\text{\ours}_V$ requires fewer output tokens, resulting in lower latency. This is because LLM inference is typically memory I/O-bound rather than computation-bound \cite{lancucki2025inference, liu2024deepseek}. These results highlight the efficiency of \ours, making it a more feasible option for practical deployment, where computational costs are a major consideration.

\section{Conclusion}
In this paper, we address the fundamental disconnect between semantic dialogue understanding and behavioral pattern modeling in CRS, which limits recommendation accuracy in LLM-based CRS. While LLMs excel at parsing explicit linguistic cues, they struggle to capture the non-semantic dependencies inherent in user behavior history. We propose \ours, a model-agnostic framework that employs latent user intent as a unifying bridge between these two representations. By formulating the recommendation process as a latent variable problem, we utilize a variational EM procedure to align textual preferences with behavioral sequences. This approach ensures that the recommendation remains dialogue-aware without necessitating the resource-intensive fine-tuning of the LLM. Our empirical evaluations across multiple real-world datasets demonstrate that \ours significantly improves recommendation accuracy in both single-turn and multi-turn dialogues and maintains high computational efficiency.
Future work may explore the dynamic evolution of latent intent over longer conversation horizons. Modeling the temporal trajectory of intent would allow the system to differentiate between persistent user preferences and transient ones, leading to more accurate recommendations and more satisfying user experiences.

\begin{acks}
This work is supported by the National Natural Science Foundation of China (No. 62576163, 62376120, 62476134).
\end{acks}

\bibliographystyle{ACM-Reference-Format}
\balance
\bibliography{sample-base}

@String{Computing = "Computing" }

@String{Springer = "Springer-Verlag" }

@ArtifactSoftware{R,
    title = {R: A Language and Environment for Statistical Computing},
    author = {{R Core Team}},
    organization = {R Foundation for Statistical Computing},
    address = {Vienna, Austria},
    year = {2019},
    url = {https://www.R-project.org/},
}

@inproceedings{jiang-etal-2024-scaling,
    title = "Scaling Sentence Embeddings with Large Language Models",
    author = "Jiang, Ting  and
      Huang, Shaohan  and
      Luan, Zhongzhi  and
      Wang, Deqing  and
      Zhuang, Fuzhen",
    editor = "Al-Onaizan, Yaser  and
      Bansal, Mohit  and
      Chen, Yun-Nung",
    booktitle = "Findings of the Association for Computational Linguistics: EMNLP 2024",
    month = nov,
    year = "2024",
}

@article{DBLP:journals/corr/abs-1807-03748,
  author       = {A{\"{a}}ron van den Oord and
                  Yazhe Li and
                  Oriol Vinyals},
  title        = {Representation Learning with Contrastive Predictive Coding},
  journal      = {CoRR},
  volume       = {abs/1807.03748},
  year         = {2018},
  url          = {http://arxiv.org/abs/1807.03748},
  eprinttype    = {arXiv},
  eprint       = {1807.03748},
}

@inproceedings{christakopoulou2018q,
  title={Q\&R: A two-stage approach toward interactive recommendation},
  author={Christakopoulou, Konstantina and Beutel, Alex and Li, Rui and Jain, Sagar and Chi, Ed H},
  booktitle={Proceedings of the 24th ACM SIGKDD international conference on knowledge discovery \& data mining},
  pages={139--148},
  year={2018}
}

@inproceedings{christakopoulou2016towards,
  title={Towards conversational recommender systems},
  author={Christakopoulou, Konstantina and Radlinski, Filip and Hofmann, Katja},
  booktitle={Proceedings of the 22nd ACM SIGKDD international conference on knowledge discovery and data mining},
  pages={815--824},
  year={2016}
}

@inproceedings{EstimationActionReflection2020,
author = {Lei, Wenqiang and He, Xiangnan and Miao, Yisong and Wu, Qingyun and Hong, Richang and Kan, Min-Yen and Chua, Tat-Seng},
title = {Estimation-Action-Reflection: Towards Deep Interaction Between Conversational and Recommender Systems},
year = {2020},
isbn = {9781450368223},
publisher = {Association for Computing Machinery},
address = {New York, NY, USA},
booktitle = {Proceedings of the 13th International Conference on Web Search and Data Mining},
pages = {304–312},
numpages = {9},
}

@article{li2018towards,
  title={Towards deep conversational recommendations},
  author={Li, Raymond and Ebrahimi Kahou, Samira and Schulz, Hannes and Michalski, Vincent and Charlin, Laurent and Pal, Chris},
  journal={Advances in neural information processing systems},
  volume={31},
  year={2018}
}

@inproceedings{zhang2018towards,
  title={Towards conversational search and recommendation: System ask, user respond},
  author={Zhang, Yongfeng and Chen, Xu and Ai, Qingyao and Yang, Liu and Croft, W Bruce},
  booktitle={Proceedings of the 27th acm international conference on information and knowledge management},
  pages={177--186},
  year={2018}
}

@inproceedings{SunZ18,
  author       = {Yueming Sun and
                  Yi Zhang},
  editor       = {Kevyn Collins{-}Thompson and
                  Qiaozhu Mei and
                  Brian D. Davison and
                  Yiqun Liu and
                  Emine Yilmaz},
  title        = {Conversational Recommender System},
  booktitle    = {The 41st International {ACM} {SIGIR} Conference on Research {\&}
                  Development in Information Retrieval, {SIGIR} 2018, Ann Arbor, MI,
                  USA, July 08-12, 2018},
  pages        = {235--244},
  publisher    = {{ACM}},
  year         = {2018},
}

@inproceedings{hu2022learning,
  title={Learning to infer user implicit preference in conversational recommendation},
  author={Hu, Chenhao and Huang, Shuhua and Zhang, Yansen and Liu, Yubao},
  booktitle={Proceedings of the 45th International ACM SIGIR conference on research and development in information retrieval},
  pages={256--266},
  year={2022}
}

@inproceedings{bi2019conversational,
  title={Conversational product search based on negative feedback},
  author={Bi, Keping and Ai, Qingyao and Zhang, Yongfeng and Croft, W Bruce},
  booktitle={Proceedings of the 28th acm international conference on information and knowledge management},
  pages={359--368},
  year={2019}
}

@inproceedings{zhang2020conversational,
  title={Conversational contextual bandit: Algorithm and application},
  author={Zhang, Xiaoying and Xie, Hong and Li, Hang and CS Lui, John},
  booktitle={Proceedings of the web conference 2020},
  pages={662--672},
  year={2020}
}

@article{jin2023lending,
  title={Lending interaction wings to recommender systems with conversational agents},
  author={Jin, Jiarui and Chen, Xianyu and Ye, Fanghua and Yang, Mengyue and Feng, Yue and Zhang, Weinan and Yu, Yong and Wang, Jun},
  journal={Advances in Neural Information Processing Systems},
  volume={36},
  pages={27951--27979},
  year={2023}
}

@article{gao2023chat,
  title={Chat-rec: Towards interactive and explainable llms-augmented recommender system},
  author={Gao, Yunfan and Sheng, Tao and Xiang, Youlin and Xiong, Yun and Wang, Haofen and Zhang, Jiawei},
  journal={arXiv preprint arXiv:2303.14524},
  year={2023}
}

@article{yang2024item,
  title={Item-Language Model for Conversational Recommendation},
  author={Yang, Li and Subbiah, Anushya and Patel, Hardik and Li, Judith Yue and Song, Yanwei and Mirghaderi, Reza and Aggarwal, Vikram},
  journal={arXiv preprint arXiv:2406.02844},
  year={2024}
}

@inproceedings{DBLP:conf/iclr/YaoZYDSN023,
  author       = {Shunyu Yao and
                  Jeffrey Zhao and
                  Dian Yu and
                  Nan Du and
                  Izhak Shafran and
                  Karthik R. Narasimhan and
                  Yuan Cao},
  title        = {ReAct: Synergizing Reasoning and Acting in Language Models},
  booktitle    = {The Eleventh International Conference on Learning Representations,
                  {ICLR} 2023, Kigali, Rwanda, May 1-5, 2023},
  publisher    = {OpenReview.net},
  year         = {2023},
}

@inproceedings{he2023large,
  title={Large language models as zero-shot conversational recommenders},
  author={He, Zhankui and Xie, Zhouhang and Jha, Rahul and Steck, Harald and Liang, Dawen and Feng, Yesu and Majumder, Bodhisattwa Prasad and Kallus, Nathan and McAuley, Julian},
  booktitle={Proceedings of the 32nd ACM international conference on information and knowledge management},
  pages={720--730},
  year={2023}
}

@article{huang2023recommender,
  title={Recommender ai agent: Integrating large language models for interactive recommendations},
  author={Huang, Xu and Lian, Jianxun and Lei, Yuxuan and Yao, Jing and Lian, Defu and Xie, Xing},
  journal={arXiv preprint arXiv:2308.16505},
  year={2023}
}

@inproceedings{yang2024unleashing,
  title={Unleashing the Retrieval Potential of Large Language Models in Conversational Recommender Systems},
  author={Yang, Ting and Chen, Li},
  booktitle={Proceedings of the 18th ACM Conference on Recommender Systems},
  pages={43--52},
  year={2024}
}

@article{he2024reindex,
  title={Reindex-Then-Adapt: Improving Large Language Models for Conversational Recommendation},
  author={He, Zhankui and Xie, Zhouhang and Steck, Harald and Liang, Dawen and Jha, Rahul and Kallus, Nathan and McAuley, Julian},
  journal={arXiv preprint arXiv:2405.12119},
  year={2024}
}

@inproceedings{kang2018self,
  title={Self-attentive sequential recommendation},
  author={Kang, Wang-Cheng and McAuley, Julian},
  booktitle={2018 IEEE international conference on data mining (ICDM)},
  pages={197--206},
  year={2018},
  organization={IEEE}
}

@inproceedings{sun2019bert4rec,
  title={BERT4Rec: Sequential recommendation with bidirectional encoder representations from transformer},
  author={Sun, Fei and Liu, Jun and Wu, Jian and Pei, Changhua and Lin, Xiao and Ou, Wenwu and Jiang, Peng},
  booktitle={Proceedings of the 28th ACM international conference on information and knowledge management},
  pages={1441--1450},
  year={2019}
}

@article{team2024gemini,
  title={Gemini 1.5: Unlocking multimodal understanding across millions of tokens of context},
  author={Team, Gemini and Georgiev, Petko and Lei, Ving Ian and Burnell, Ryan and Bai, Libin and Gulati, Anmol and Tanzer, Garrett and Vincent, Damien and Pan, Zhufeng and Wang, Shibo and others},
  journal={arXiv preprint arXiv:2403.05530},
  year={2024}
}

@article{liu2023chatgpt,
  title={Is chatgpt a good recommender? a preliminary study},
  author={Liu, Junling and Liu, Chao and Zhou, Peilin and Lv, Renjie and Zhou, Kang and Zhang, Yan},
  journal={arXiv preprint arXiv:2304.10149},
  year={2023}
}

@inproceedings{cen2020controllable,
  title={Controllable multi-interest framework for recommendation},
  author={Cen, Yukuo and Zhang, Jianwei and Zou, Xu and Zhou, Chang and Yang, Hongxia and Tang, Jie},
  booktitle={Proceedings of the 26th ACM SIGKDD International Conference on Knowledge Discovery \& Data Mining},
  pages={2942--2951},
  year={2020}
}

@inproceedings{li2019multi,
  title={Multi-interest network with dynamic routing for recommendation at Tmall},
  author={Li, Chao and Liu, Zhiyuan and Wu, Mengmeng and Xu, Yuchi and Zhao, Huan and Huang, Pipei and Kang, Guoliang and Chen, Qiwei and Li, Wei and Lee, Dik Lun},
  booktitle={Proceedings of the 28th ACM international conference on information and knowledge management},
  pages={2615--2623},
  year={2019}
}

@article{li2021intention,
  title={Intention-aware sequential recommendation with structured intent transition},
  author={Li, Haoyang and Wang, Xin and Zhang, Ziwei and Ma, Jianxin and Cui, Peng and Zhu, Wenwu},
  journal={IEEE Transactions on Knowledge and Data Engineering},
  volume={34},
  number={11},
  pages={5403--5414},
  year={2021},
  publisher={IEEE}
}

@inproceedings{liu2020basket,
  title={Basket recommendation with multi-intent translation graph neural network},
  author={Liu, Zhiwei and Li, Xiaohan and Fan, Ziwei and Guo, Stephen and Achan, Kannan and Philip, S Yu},
  booktitle={2020 IEEE International Conference on Big Data (Big Data)},
  pages={728--737},
  year={2020},
  organization={IEEE}
}

@inproceedings{pan2020intent,
  title={An intent-guided collaborative machine for session-based recommendation},
  author={Pan, Zhiqiang and Cai, Fei and Ling, Yanxiang and de Rijke, Maarten},
  booktitle={Proceedings of the 43rd international ACM SIGIR conference on research and development in information retrieval},
  pages={1833--1836},
  year={2020}
}

@inproceedings{tanjim2020attentive,
  title={Attentive sequential models of latent intent for next item recommendation},
  author={Tanjim, Md Mehrab and Su, Congzhe and Benjamin, Ethan and Hu, Diane and Hong, Liangjie and McAuley, Julian},
  booktitle={Proceedings of The Web Conference 2020},
  pages={2528--2534},
  year={2020}
}

@inproceedings{chen2022intent,
  title={Intent contrastive learning for sequential recommendation},
  author={Chen, Yongjun and Liu, Zhiwei and Li, Jia and McAuley, Julian and Xiong, Caiming},
  booktitle={Proceedings of the ACM Web Conference 2022},
  pages={2172--2182},
  year={2022}
}

@inproceedings{sun2024large,
  title={Large language models for intent-driven session recommendations},
  author={Sun, Zhu and Liu, Hongyang and Qu, Xinghua and Feng, Kaidong and Wang, Yan and Ong, Yew Soon},
  booktitle={Proceedings of the 47th International ACM SIGIR Conference on Research and Development in Information Retrieval},
  pages={324--334},
  year={2024}
}

@inproceedings{zhang2024usimagent,
  title={Usimagent: Large language models for simulating search users},
  author={Zhang, Erhan and Wang, Xingzhu and Gong, Peiyuan and Lin, Yankai and Mao, Jiaxin},
  booktitle={Proceedings of the 47th International ACM SIGIR Conference on Research and Development in Information Retrieval},
  pages={2687--2692},
  year={2024}
}

@inproceedings{yoon2024evaluating,
  title={Evaluating Large Language Models as Generative User Simulators for Conversational Recommendation},
  author={Yoon, Se-eun and He, Zhankui and Echterhoff, Jessica and McAuley, Julian},
  booktitle={Proceedings of the 2024 Conference of the North American Chapter of the Association for Computational Linguistics: Human Language Technologies (Volume 1: Long Papers)},
  pages={1490--1504},
  year={2024}
}

@article{hou2024bridging,
  title={Bridging Language and Items for Retrieval and Recommendation},
  author={Hou, Yupeng and Li, Jiacheng and He, Zhankui and Yan, An and Chen, Xiusi and McAuley, Julian},
  journal={arXiv preprint arXiv:2403.03952},
  year={2024}
}

@inproceedings{deng2021unified,
  title={Unified conversational recommendation policy learning via graph-based reinforcement learning},
  author={Deng, Yang and Li, Yaliang and Sun, Fei and Ding, Bolin and Lam, Wai},
  booktitle={Proceedings of the 44th International ACM SIGIR Conference on Research and Development in Information Retrieval},
  pages={1431--1441},
  year={2021}
}

@article{dubey2024llama,
  title={The llama 3 herd of models},
  author={Dubey, Abhimanyu and Jauhri, Abhinav and Pandey, Abhinav and Kadian, Abhishek and Al-Dahle, Ahmad and Letman, Aiesha and Mathur, Akhil and Schelten, Alan and Yang, Amy and Fan, Angela and others},
  journal={arXiv preprint arXiv:2407.21783},
  year={2024}
}

@article{achiam2023gpt,
  title={Gpt-4 technical report},
  author={Achiam, Josh and Adler, Steven and Agarwal, Sandhini and Ahmad, Lama and Akkaya, Ilge and Aleman, Florencia Leoni and Almeida, Diogo and Altenschmidt, Janko and Altman, Sam and Anadkat, Shyamal and others},
  journal={arXiv preprint arXiv:2303.08774},
  year={2023}
}

@article{wei2022chain,
  title={Chain-of-thought prompting elicits reasoning in large language models},
  author={Wei, Jason and Wang, Xuezhi and Schuurmans, Dale and Bosma, Maarten and Xia, Fei and Chi, Ed and Le, Quoc V and Zhou, Denny and others},
  journal={Advances in neural information processing systems},
  volume={35},
  pages={24824--24837},
  year={2022}
}

@article{luo2023wizardcoder,
  title={Wizardcoder: Empowering code large language models with evol-instruct},
  author={Luo, Ziyang and Xu, Can and Zhao, Pu and Sun, Qingfeng and Geng, Xiubo and Hu, Wenxiang and Tao, Chongyang and Ma, Jing and Lin, Qingwei and Jiang, Daxin},
  journal={arXiv preprint arXiv:2306.08568},
  year={2023}
}

@article{farshidi2024understanding,
  title={Understanding user intent modeling for conversational recommender systems: a systematic literature review},
  author={Farshidi, Siamak and Rezaee, Kiyan and Mazaheri, Sara and Rahimi, Amir Hossein and Dadashzadeh, Ali and Ziabakhsh, Morteza and Eskandari, Sadegh and Jansen, Slinger},
  journal={User Modeling and User-Adapted Interaction},
  pages={1--64},
  year={2024},
  publisher={Springer}
}

@inproceedings{xi2024memocrs,
  title={MemoCRS: Memory-enhanced Sequential Conversational Recommender Systems with Large Language Models},
  author={Xi, Yunjia and Liu, Weiwen and Lin, Jianghao and Chen, Bo and Tang, Ruiming and Zhang, Weinan and Yu, Yong},
  booktitle={Proceedings of the 33rd ACM International Conference on Information and Knowledge Management},
  pages={2585--2595},
  year={2024}
}

@inproceedings{ravaut2024parameter,
  title={Parameter-Efficient Conversational Recommender System as a Language Processing Task},
  author={Ravaut, Mathieu and Zhang, Hao and Xu, Lu and Sun, Aixin and Liu, Yong},
  booktitle={Proceedings of the 18th Conference of the European Chapter of the Association for Computational Linguistics (Volume 1: Long Papers)},
  pages={152--165},
  year={2024}
}

@inproceedings{bao2024large,
  title={Large language models for recommendation: Past, present, and future},
  author={Bao, Keqin and Zhang, Jizhi and Lin, Xinyu and Zhang, Yang and Wang, Wenjie and Feng, Fuli},
  booktitle={Proceedings of the 47th International ACM SIGIR Conference on Research and Development in Information Retrieval},
  pages={2993--2996},
  year={2024}
}

@inproceedings{kemper2024retrieval,
  title={Retrieval-Augmented Conversational Recommendation with Prompt-based Semi-Structured Natural Language State Tracking},
  author={Kemper, Sara and Cui, Justin and Dicarlantonio, Kai and Lin, Kathy and Tang, Danjie and Korikov, Anton and Sanner, Scott},
  booktitle={Proceedings of the 47th International ACM SIGIR Conference on Research and Development in Information Retrieval},
  pages={2786--2790},
  year={2024}
}

@article{siro2023understanding,
  title={Understanding and predicting user satisfaction with conversational recommender systems},
  author={Siro, Clemencia and Aliannejadi, Mohammad and De Rijke, Maarten},
  journal={ACM Transactions on Information Systems},
  volume={42},
  number={2},
  pages={1--37},
  year={2023},
  publisher={ACM New York, NY}
}

@inproceedings{mahmud2025understanding,
  title={Understanding user preferences for interaction styles in conversational recommender system: The predictive role of system qualities, user experience, and traits},
  author={Mahmud, Raj and Berkovsky, Shlomo and Prasad, Mukesh and Kocaballi, A Baki},
  booktitle={Proceedings of the 37th Australian Conference on Human-Computer Interaction},
  pages={68--80},
  year={2025}
}

@article{lancucki2025inference,
  title={Inference-Time Hyper-Scaling with KV Cache Compression},
  author={{\L}a{\'n}cucki, Adrian and Staniszewski, Konrad and Nawrot, Piotr and Ponti, Edoardo M},
  journal={arXiv preprint arXiv:2506.05345},
  year={2025}
}

@article{liu2024deepseek,
  title={Deepseek-v3 technical report},
  author={Liu, Aixin and Feng, Bei and Xue, Bing and Wang, Bingxuan and Wu, Bochao and Lu, Chengda and Zhao, Chenggang and Deng, Chengqi and Zhang, Chenyu and Ruan, Chong and others},
  journal={arXiv preprint arXiv:2412.19437},
  year={2024}
}

@inproceedings{wang2022towards,
  title={Towards unified conversational recommender systems via knowledge-enhanced prompt learning},
  author={Wang, Xiaolei and Zhou, Kun and Wen, Ji-Rong and Zhao, Wayne Xin},
  booktitle={Proceedings of the 28th ACM SIGKDD conference on knowledge discovery and data mining},
  pages={1929--1937},
  year={2022}
}

@inproceedings{zhu2025collaborative,
  title={Collaborative retrieval for large language model-based conversational recommender systems},
  author={Zhu, Yaochen and Wan, Chao and Steck, Harald and Liang, Dawen and Feng, Yesu and Kallus, Nathan and Li, Jundong},
  booktitle={Proceedings of the ACM on Web Conference 2025},
  pages={3323--3334},
  year={2025}
}

\appendix

\section{Proof of the Lower Bound}\label{abb:variations_prove}
In this section, we prove that the ELBO is a lower bound on $\log p(v \mid \mathcal{H}^u, x^u)$.
\begin{equation}
\begin{aligned} 
\log & p(v \mid \mathcal{H}^u, x^u) = \log \sum_{m_j \in \mathcal{M}} p(v, m_j \mid \mathcal{H}^u, x^u) \\ 
= \log &\sum_{m_j \in \mathcal{M}} q(m_j \mid \mathcal{H}^u) \frac{p(v, m_j \mid \mathcal{H}^u, x^u)}{q(m_j \mid \mathcal{H}^u)} \\ 
\ge &\sum_{m_j \in \mathcal{M}} q(m_j \mid \mathcal{H}^u) \log \frac{p(v, m_j \mid \mathcal{H}^u, x^u)}{q(m_j \mid \mathcal{H}^u)} \\ 
= &\sum_{m_j \in \mathcal{M}} q(m_j \mid \mathcal{H}^u) \log \frac{p(v \mid m_j, \mathcal{H}^u, x^u) p(m_j \mid \mathcal{H}^u, x^u)}{q(m_j \mid \mathcal{H}^u)} \\ 
= &\sum_{m_j \in \mathcal{M}} q(m_j \mid \mathcal{H}^u) \Bigl( \log p(v \mid m_j, \mathcal{H}^u, x^u) + \log p(m_j \mid \mathcal{H}^u, x^u) \\ & - \log q(m_j \mid \mathcal{H}^u) \Bigr) \\ 
= &\text{ELBO} 
\end{aligned}
\end{equation}

\section{Proof of M-step Optimization Loss}\label{abb:infonce_prove}
The derivation of the loss function $\mathcal{L}_M$ begins by decomposing the Evidence Lower Bound (ELBO) into a reconstruction term and a regularization term. The reconstruction term, which represents the log-likelihood of identifying the correct item, is computationally intractable to optimize directly. We therefore substitute it with the InfoNCE objective, which serves as a lower bound on the mutual information and an effective proxy for the log-likelihood via negative sampling. Minimizing the negative ELBO is thus equivalent to minimizing the combined InfoNCE loss and the Kullback-Leibler (KL) divergence. 
\begin{equation}
\begin{aligned}
\text{ELBO} &= \sum_{m \in \mathcal{M}} q(m \mid \mathcal{H}^u) \left( \log p(v \mid m, \mathcal{H}^u, x^u)  + \log p(m \mid \mathcal{H}^u, x^u) \right. \\ & \qquad \left. - \log q(m \mid \mathcal{H}^u) \right) \\
& = \sum_{m \in \mathcal{M}} q(m \mid \mathcal{H}^u) \log p(v \mid m, \mathcal{H}^u, x^u) \\ & \qquad - \sum_{m \in \mathcal{M}} q(m \mid \mathcal{H}^u) \log \frac{q(m \mid \mathcal{H}^u)}{p(m \mid \mathcal{H}^u, x^u)} \\
& = \mathbb{E}_{q(m \mid \mathcal{H}^u)} \bigl[\log p(v \mid m, \mathcal{H}^u, x^u)\bigr] \\ & \qquad - \text{KL}\bigl(q(m \mid \mathcal{H}^u) \,\|\, p(m \mid \mathcal{H}^u, x^u)\bigr)
\end{aligned}
\end{equation}
To maximize the ELBO, we minimize its negative, defined as the loss $\mathcal{L}$. We introduce a weighting parameter $\lambda$ for the KL term:
\begin{equation}
\begin{aligned}
    \mathcal{L} = &-\mathbb{E}_{q(m \mid \mathcal{H}^u)} \bigl[\log p(v \mid m, \mathcal{H}^u, x^u)\bigr] + \\ & \qquad \lambda \, \text{KL}\bigl(q(m \mid \mathcal{H}^u) \,\|\, p(m \mid \mathcal{H}^u, x^u)\bigr)
\end{aligned}
\end{equation}
The negative log-likelihood is approximated using the InfoNCE loss $\ell_{NCE}$ with negative sampling:
\begin{equation}
\begin{aligned}
    -\log p(v \mid m, \mathcal{H}^u, x^u) & \approx \ell_{NCE}(v, m) \\ & = - \log \frac{\exp(s(v, m, \mathcal{H}^u, x^u))}{\sum_{v' \in \{v\} \cup \mathcal{V}_{neg}} \exp(s(v', m, \mathcal{H}^u, x^u))}
\end{aligned}
\end{equation}
Substituting this approximation into the global loss over all user-item pairs $(u,v)$ gives:
\begin{equation}
\begin{aligned}
    \mathcal{L}_M &= \sum_{(u,v)} \left( \sum_{m \in \mathcal{M}} q(m \mid \mathcal{H}^u) \, \ell_{NCE}(v, m) \right) \\ & \qquad + \lambda \, \text{KL}\bigl(q(m \mid \mathcal{H}^u) \,\|\, p(m \mid \mathcal{H}^u, x^u)\bigr) \\
    &= \mathcal{L}_{NCE} + \lambda \, \text{KL}\bigl(q(m \mid \mathcal{H}^u) \,\|\, p(m \mid \mathcal{H}^u, x^u)\bigr)
\end{aligned}
\end{equation}

Here, optimizing this InfoNCE objective is theoretically grounded. The InfoNCE loss maximizes a lower bound on the mutual information between the context and the target item. Furthermore, as the number of negative samples $|\mathcal{V}_{neg}|$ increases, InfoNCE provides a tractable contrastive surrogate for the intractable likelihood term $\log p_\theta(v \mid m, \dots)$, thereby justifying its use as a tractable proxy for our probabilistic objective.

\begin{table*}[t]\small
  \centering
  \caption{Efficiency of different methods. Inference time is measured only for locally deployed LLMs. For commercial black-box models, we report token counts without measuring runtime.}
  \resizebox{0.6\linewidth}{!}{
    \begin{tabular}{l|ccc|ccc|ccc}
    \toprule
          & \multicolumn{3}{c|}{MovieLens} & \multicolumn{3}{c|}{VideoGames} & \multicolumn{3}{c}{CDs} \\
          & Input & Output & Time(ms) & Input & Output & Time(ms) & Input & Output & Time(ms) \\
    \midrule
    Llama-3.1-8B &   119.51    &  148.20  &  2568 &   127.20   &   199.83    & 2635 &  124.26   &  211.57 & 3605 \\
    Gemini-1.5-Pro &   120.49    &  140.56    &- &   112.43  &   188.75    & -&  126.89   &  223.91 & -\\
    GPT-4  &    118.95   &    152.81   &- &   106.85    &   181.36    &- &    129.88   &  235.46 & -\\
    \midrule
    $\text{\ours}_B$ & 85.50 & 1.00 & 87 & 75.44 & 1.00 & 89 & 91.90 & 1.00 & 92 \\
    $\text{\ours}_F$ & 317.50 & 35.59 & 709 & 781.44 & 50.76 & 972 & 2086.89 & 70.70 & 1284 \\
    $\text{\ours}_V$ & 810.26 & 98.63 & 2168 & 2117.60  & 96.72 & 2489 & 2178.55 & 111.98 & 2488 \\
    \bottomrule
    \end{tabular}}
  \label{app:cost}%
\end{table*}%

\begin{table}[h]
  \centering
  \caption{Dataset statistics. Avg. denotes average.}
  \resizebox{\linewidth}{!}{
    \begin{tabular}{l|ccc}
    \toprule
          & MovieLens-1M & VideoGames & CDs and Vinyl \\
    \midrule
    Number of users & 6034  & 15582 & 104544 \\
    Number of items & 3522  & 7233  & 76616 \\
    Actions & 575272 & 122179 & 1259069 \\
    Avg. behavior length & 95.34 & 7.84  & 12.04 \\
    Sparsity & 97.30\% & 99.89\% & 99.98\% \\
    \bottomrule
    \end{tabular}
    }
  \label{datasets}
\end{table}
\section{Assistant Recommendation Module}\label{abb_assrec}
Relying solely on the KL signal can be unstable, particularly in the early stages of training when the generative model is not yet mature. To address this, we introduce an auxiliary recommendation objective. We compute a soft intent representation $\mathbf{r}^u$ by pooling the anchors according to the predicted distribution:
\begin{equation}
\mathbf{r}^u = \sum_{m_k \in \mathcal{M}} q_\phi(m_k \mid \mathcal{H}^u) \cdot \mathbf{m}_k
\end{equation}
We also apply an InfoNCE loss to ensure this intent vector $\mathbf{r}^u$ is capable of retrieving the ground-truth item $v$:
\begin{equation}\label{E_rec}
\mathcal{L}^{rec} = - \sum_{(u,v) \in \mathcal{D}} \log \frac{\exp(\mathbf{r}^u \cdot \mathbf{v} / \tau)}{\exp(\mathbf{r}^u \cdot \mathbf{v} / \tau) + \sum_{v^- \in \mathcal{V}_{neg}} \exp(\mathbf{r}^u \cdot \mathbf{v}^- / \tau)}
\end{equation}
where $\mathcal{V}_{neg}$ is a set of negative samples.

\section{Experimental Settings}
\subsection{Datasets}\label{abb:dataset}
Each dataset contains user-item interaction history and item metadata. Since not all items in the Amazon VideoGames and Amazon CDs datasets have associated metadata, we preprocess the raw data to exclude items lacking metadata. Additionally, we focus on the ``5-core'' data, where each user and item has at least five interactions.
The key statistics of the datasets are summarized in Table \ref{datasets} in Appendix. We follow the data splitting methodology outlined in \cite{kang2018self, sun2019bert4rec, chen2022intent}. Specifically, we use the last item in each user's sequence as the test set, the second-to-last item as the validation set, and all other items as the training set.

\section{Additional Experimental Results}
\subsection{Implementation Details} \label{app_impledetails}
For the recommendation component, we implement the ICLRec \cite{chen2022intent} model. All optimization steps are carried out using PyTorch. Following \cite{gao2023chat}, we limit the maximum number of interaction turns to 5 and present at most 5 items in each dialogue turn.

Moreover, as shown in Section \ref{intentrepresentation}, a single user's comprehensive history encapsulates multiple distinct latent intents, each corresponding to a specific behavioral context or session. To fully capture this diversity and prevent the model from overfitting to a static user profile, we employ a random sequence segmentation strategy. By slicing user behavior sequences into multiple subsequences, we treat each segment as an independent interaction instance. This approach not only artificially expands the dataset but also simulates a broader spectrum of partial history contexts. Consequently, it forces the model to learn robust mappings between varying behavioral snapshots and their corresponding latent intent anchors ($m_k$), thereby enhancing generalization across diverse interaction scenarios.

For the semantic representation \(\mathbf{x}^u\) of user textual descriptions $x^u$, we follow recent studies that obtain embeddings from LLMs~\cite{jiang-etal-2024-scaling}, using the following prompt: \emph{``This sentence: `*sentence*' means in one word:''} to extract a semantic embedding. Furthermore, we implement a semantic caching scheme to optimize training efficiency. Since the linguistic content of the dialogue history remains static during the optimization of the recommendation module, we pre-compute and store the embeddings for all natural language descriptions. This strategy offloads the computationally intensive LLM inference from the training loop, significantly reducing overall training time.

\subsection{Efficiency}
Here, we demonstrate the efficiency of \ours across all three datasets, as shown in Table \ref{app:cost}. Given that the parameters of LLMs are significantly larger than those of the recommendation model, generative model, and inference model, most of the latency is caused by LLM inference. There is a near-linear relationship between the number of LLM output tokens and inference time. Furthermore, the input tokens do not have a substantial impact on the time cost, as modern computational hardware makes LLM inference memory I/O-bound rather than computation-bound.

\section{Prompts for User Simulator} \label{abb:prompts}
You are a user chatting with a recommender for \{item\} recommendation in a multi-turn conversation.
Your history is \{history\}. Your target item is \{target\}.
Here is the information about the target item that you may use: \{target item info\} \\
You must follow the instructions below during the chat. \\
State your preferences to guide the recommender toward the target item. \\
If the recommender recommends \{target\}, accept it. \\
If the recommender recommends other items, refuse them and provide information about the target item. \\
Never directly reveal the target item title. \\
If the recommender asks about your preferences, provide information about the target item. \\
You may provide your history. \\
Your output should contain only the user's utterance. \\
If you think the conversation has ended, output \texttt{<END>}. \\
Do not provide too many details about the target item at once; fewer than three conditions is better. \\
Now start first and act as the user.
\end{document}